%% file: camera-ready.tex
\pdfoutput=1
\documentclass[11pt]{article}

\usepackage{acl}
\usepackage{times}
\usepackage{latexsym}

\usepackage[T1]{fontenc}
\usepackage[utf8]{inputenc}
\usepackage{amsmath, amssymb}
\usepackage{mathtools}

\usepackage{microtype}
\usepackage{multirow}
\usepackage{caption}
\usepackage{makecell}
\usepackage{arydshln}
\usepackage{array,multirow,graphicx}
\usepackage{float}
\PassOptionsToPackage{dvipsnames,svgnames,table}{xcolor}

\usepackage{fixltx2e}
\usepackage{fontawesome}
\usepackage{svg}

\def\SPSB#1#2{\rlap{\textsuperscript{#1}}\SB{#2}}
\def\SP#1{\textsuperscript{#1}}
\def\SB#1{\textsubscript{#1}}

\definecolor{purplee}{HTML}{AE2FEB}
\definecolor{pink}{HTML}{F477AD}
\definecolor{gptcolor}{HTML}{22C99A}
\definecolor{xlmcolor}{HTML}{883BEB}
\definecolor{lstmcolor}{HTML}{F45D51}
\definecolor{mlpcolor}{HTML}{2796FE}
\definecolor{gcncolor}{HTML}{0B8F88}
\definecolor{blue}{HTML}{1C70D6}
\definecolor{green}{HTML}{1F945D}
\definecolor{red}{HTML}{D64531}

\newcommand{\bloom}{\textcolor{pink}{\faGe}}
\newcommand{\mgpt}{\textcolor{gptcolor}{\faGg}}
\newcommand{\lstm}{\textcolor{lstmcolor}{$\to$}}
\newcommand{\bilstm}{\textcolor{lstmcolor}{$\leftrightarrow$}}
\newcommand{\xlm}{\textcolor{xlmcolor}{\faSkyatlas}}

\newcommand{\abs}{A}
\newcommand{\rel}{R}
\newcommand{\relpos}{P}
\newcommand{\pos}{P}
\newcommand{\planari}{1P}
\newcommand{\planarii}{2P}

\definecolor{flcolor}{HTML}{1F77B4}
\definecolor{nonflcolor}{HTML}{FF7F03}
\definecolor{tbcolor}{HTML}{2CA02C}
\newcommand{\fl}{\textcolor{flcolor}{\small\faSquare}}
\newcommand{\nonfl}{\textcolor{nonflcolor}{\small\faCircleO}}
\newcommand{\tb}{\textcolor{tbcolor}{\small\faStarO}}

\title{On the Challenges of Fully Incremental Neural Dependency Parsing}

\author
{
    Ana Ezquerro, Carlos Gómez-Rodríguez and David Vilares\\
    Universidade da Coru\~{n}a, CITIC \\
    Departamento de Ciencias de la Computación y Tecnologías de la Información \\
    Campus de Elvi\~{n}a s/n, 15071 \\ A Coru\~{n}a, Spain \\
    \texttt{\{ana.ezquerro, carlos.gomez, david.vilares\}@udc.es} \\
}

\begin{document}
\maketitle
\begin{abstract}
Since the popularization of BiLSTMs and Transformer-based bidirectional encoders, state-of-the-art syntactic parsers have lacked incrementality, requiring access to the whole sentence and deviating from human language processing. This paper explores whether fully incremental dependency parsing with modern architectures can be competitive. We build parsers combining strictly left-to-right neural encoders with fully incremental sequence-labeling and transition-based decoders.  The results show that fully incremental parsing with modern architectures considerably lags behind bidirectional parsing, noting the challenges of psycholinguistically plausible parsing.
\end{abstract}

\section{Introduction}\label{introduction}

Human understanding of natural language is widely agreed to be \emph{incremental}: humans do not need to read a complete sentence to start understanding it. Instead, we update partial interpretations as we receive more input \citep{MarslenWilson85}. 

While the exact way in which this incrementality works is still unclear~\cite{kitaev-etal-2022-learned}, its presence implies that some form of incrementality is an obvious necessary condition for a parser to be psycholinguistically plausible as a model of human processing \citep{miller-schuler-2010-hhmm}. Since human processing is the gold standard for automatic parsing, we know that it should be possible to achieve accurate parsing with incremental systems. Yet, in recent years, none of the competitive syntactic parsers that have been proposed for either of the main syntactic formalisms can be said to be incremental, even under the loosest possible definitions of the term. This poses challenges in the intersection between syntax and computational psycholinguistics, e.g., use cases both for modeling of human parsing and for real-time settings where one wants partial results before waiting for a sentence to end. Currently, most parsers use bidirectional encoders, such as BiLSTMs \citep{kiperwasser-goldberg-2016-simple,Biaffine} or Transformers \citep{zhou-zhao-2019-head,mrini-etal-2020-rethinking,AttachJuxtapose}, so the whole sentence is being used before even processing the first word. An exception is the constituent parser
by \citet{kitaev-etal-2022-learned}, who use a fully incremental encoder, but the rest of the model is bidirectional, as it uses Transformer layers and a CYK-like, non-left-to-right span-based decoder~\citep{stern-etal-2017-minimal}.

This paper explores the viability of \emph{fully incremental} dependency parsing, i.e., parsers where all the components (from the encoder to the decoder) work \emph{strictly} from left to right. To our knowledge, this is the first attempt to build fully incremental dependency parsers with modern deep learning architectures.

\section{Incrementality in Parsing}
\paragraph{In transition-based parsing} 
Transition-based parsing has traditionally been linked to incrementality~\citep{nivre-2008-algorithms}, as it works from left to right and builds partial outputs. Some authors consider that transition-based parsers as a whole are incremental, as they have internal states with partial outputs~\citep{IncrementalParserStates}. We will call this criterion \emph{weak incrementality}. Others exclude algorithms like the arc-standard dependency parser, where dependencies are not built in left-to-right order and input arbitrarily far in the future might be needed to build right-branching dependencies~\citep{christiansen_chater_2016}. We will call this stricter view \emph{strong incrementality}, and formalize it as follows: given a monotonic parser (i.e., one where each partial parse is a superset of the previous), we say that it is \emph{strongly incremental with delay} $k$ if every possible partial parse for a prefix $w_1 \ldots w_{i-k}$ can be built upon reading the prefix $w_1 \ldots w_i$, without the parser having accessed the rest of the input.\footnote{The definition of a partial parse for the prefix $w_1 \ldots w_{i-k}$ depends on the grammatical formalism. For dependency parsing, we mean the subgraph of a full parse induced by the nodes in $w_1 \ldots w_{i-k}$. Some authors require connectedness~\citep{Beuck11incremental}. However, since the path between two words in $w_1 \ldots w_{i-k}$ in the final parse may involve words outside the prefix, we do not believe this requirement is necessary.} Analogous considerations about the limitations with right-branching in weak incrementality, and parsers that try to avoid it to various extents, have been studied in the CCG literature~\citep{ambati-etal-2015-incremental,stanojevic-steedman-2019-ccg,stanojevic-steedman-2020-max,stanojevic-etal-2021-modeling}. Contrary to arc-standard, other transition-based dependency parsers are strongly incremental: the arc-eager~\citep{nivre-2003-efficient}, Covington~\citep{covington01} or multiplanar~\citep{gomez-rodriguez-nivre-2013-divisible} parsers all fit our definition above.

Classic implementations of strongly incremental parsers typically have positive delay~\citep{Beuck11incremental} between input and output due to lookahead. Some approaches have considered zero delay, albeit with weaker performance~\citep{kohn-menzel-2013-incremental}. Solutions are also available for speculativity in incremental parsing~\citep{kitaev-etal-2022-learned}, by introducing non-monotonicity \citep{honnibal-etal-2013-non,fernandez-gonzalez-gomez-rodriguez-2017-full}.

Thus, the paradigm supports incrementality and many implementations of these parsers from the pre-deep-learning era, which did not use contextualized encoders, were strongly incremental, leading to the observation by \citet{GomBBS2016} that at that point, some state-of-the-art parsing models were converging with psycholiguistically plausible models. However, in recent years bidirectional encoders have become ubiquitous, ruling out even weak incrementality from recent implementations of transition-based parsers, be them for dependency or other grammatical formalisms \citep{kiperwasser-goldberg-2016-simple,stanojevic-steedman-2019-ccg,fernandez-astudillo-etal-2020-transition,HierarchicalPN}. In this respect, it is worth mentioning that the  approach by \citet{AttachJuxtapose} is described as ``strongly incremental constituency parsing'' but this refers to the decoder, as they use a bidirectional encoder. The only recent proposal we are aware of that aims for incrementality in the whole system is the CCG parser by \citet{stanojevic-steedman-2020-max}, also a constituency parser, but its labelled F-score is over 7 points lower than a non-incremental baseline in an English-only evaluation.

\paragraph{In label-based parsing} Other parsing paradigms that yield themselves to incrementality, as they could work from left to right, are seq2seq parsing~\citep{Vinyals2015} and sequence-labeling parsing~\citep{gomez-rodriguez-vilares-2018-constituent,strzyz-etal-2019-viable}. However, for the former, we are not aware of any implementation without bidirectional encoders. For the latter, while there are strongly incremental sequence-labeling decoders for both constituency~\citep{gomez-rodriguez-vilares-2018-constituent} and dependency~\citep{strzyz-etal-2020-bracketing}, most implementations use bidirectional encoders as well. The exception are some experiments with feed-forward encoders in \citet{gomez-rodriguez-vilares-2018-constituent}, using a sliding window to model near future context (and thus, with delay). Yet, their F-score is 14 points below their non-incremental counterparts in the same paper, and almost 20 below the overall state of the art.

\section{Incremental models}
\label{sec:models}

The research question arises whether it is possible to have competitive incremental dependency parsers in the neural era. We take the first step and test how mainstream approaches would work in a setting of strong incrementality. In our work, we will focus on models with \emph{strictly} zero delay, but we also evaluate less strict setups, in particular with delays 1 and 2. To do so, we will rely on modern encoder-decoder models. All source code is available on GitHub (\url{https://github.com/anaezquerro/incpar}).

\subsection{Incremental encoders}

Let $w=[w_1,w_2,...w_{|w|}]$ (with $w_i \in \mathcal{V}$) be an input sentence. An encoder can be seen as a parameterized function $\Omega_{\theta,|w|}$ :$ \mathcal{V}^{|w|} \rightarrow \mathcal{H}^{|w|}$, where $\mathcal{V}$ is the input vocabulary space, and $\mathcal{H} \in \mathcal{R}^N$ is the hidden representational space in where each $w_i$ is projected. In this work we are particularly interested in incremental encoders, i.e., those where given a token $w_i$, the computation of its projected representation $h_i$ only needs the sub-sequence $w_{[1:i]}$. We consider different encoders for this purpose: (i) \textbf{4 stacked left-to-right LSTMs}~\cite{hochreiter1997long}, where input is a concatenation of a word and PoS tag vector (random init) and a char-level unidirectional LSTM; (ii) \textbf{BLOOM} \cite{scao2022bloom} (due to resource constraints, we run the smallest version with 560M parameters); and (iii) \textbf{mGPT} \cite{mgpt}.

\noindent As \emph{control} encoders (upper bound baselines), we use non-incremental encoders:  (i) bidirectional LSTMs (same setup as for left-to-right LSTMs),  and (ii) XLM-RoBERTa \cite{conneau-etal-2020-unsupervised}.

\subsection{Incremental decoders}

We consider incremental (i) sequence labeling parsing, and (ii) transition-based parsing decoders.

\subsubsection{Sequence labeling decoders}\label{section-decoders-sl}

A sequence labeling decoder is a parametrized function $\Phi_{\theta,|w|}$: $\mathcal{H}^{|w|} \rightarrow \mathcal{L}^{|w|}$, which maps  each hidden vector ($h_i \in \mathcal{H}$) outputted by a generic encoder into an output label $l_i \in \mathcal{L}$ that represents a part of the output parse.  As the decoder, we use a 1-layered feed-forward network and a softmax. As for label encodings, we select representatives from two encoding families \cite{strzyz-etal-2019-viable,strzyz-etal-2020-bracketing}:

\medskip
 
\noindent\textbf{Head-based} We study three variants, all of them supporting non-projective trees. First, the absolute-indexing encoding (abs-idx), where the token labels are the index of their head. Second, the relative-indexing encoding (rel-idx),
where the label is the difference between the head and dependent indexes. Third, the PoS-tag-based encoding (PoS-idx), where each label is encoded as an offset that indicates that the $n$th word to its left/right with a given PoS tag is the head.\footnote{The PoS-tag based encoding needs PoS tags for decoding the sequence of labels to a tree. Instead of introducing PoS-tag information in those models, our PoS-tag-based decoders predict in multitask learning both the syntactic label and PoS tag associated to each word, in order to remove bias with respect to other encodings.}

\medskip

\noindent\textbf{Strings of brackets} 
First, we consider the 1-planar bracketing encoding (1p), where the label for each token is represented using a string of brackets, with each arc represented by a bracket pair. This encoding can only model crossing arcs in opposite directions. To tackle this, there is a 2-planar variant (2p), analogous, but defining a second plane of brackets.

\medskip
 
\noindent In the context of full incrementality, we will say that an encoding is \emph{forward-looking} if a label for a token $w_i$ can refer to some token to the right of $w_i$. The abs-idx, rel-idx and PoS-idx encodings are forward-looking (e.g., with abs-idx, the word $w_2$ could have $4$ as its label, meaning that its head is $w_4$, which has not been read yet); while the bracketing encodings are \emph{not} forward-looking. Forward-lookingness does not break  incrementality: all the considered encodings are still strongly incremental with delay 0 (all dependencies involving $w_1 \ldots w_i$ can be retrieved from the labels $l_1 \ldots l_i$). However, one could expect forward-looking encodings to suffer more from using incremental encoders, due to needing to make decisions involving future words that the system cannot yet access.

In our implementation, for models with delay zero, the $i$th label is predicted directly from $h_i$. For models with delay $k>0$, labels are predicted from a concatenated representation $h_i \cdot ... \cdot h_{i+k-1} \cdot h_{i+k}$.

It is also worth noting that the obtention of the tree encoded by a sequence labeling encoding can require simple postprocessing heuristics (e.g. to remove cycles in head-selection encodings). This does not break incrementality, as these heuristics are applicable to partial outputs as well.

\subsubsection{Transition-based decoders}\label{section-decoders-tb}

A transition-based decoder is defined as a tuple ($C$, $T$, c$_s$, C$_t$), where $C$ is a set of configurations (or parsing states) with associated partial parses, $T$ a set of transitions between states, and $c_s$ and $C_t$ are the initial state and set of valid final states, respectively. In the case of the arc-eager parser~\cite{nivre-2008-algorithms}, states are triplets of the form ($\sigma$,$\beta$,$A$) where $\sigma$ is a stack of partially processed words, $\beta$ a buffer of remaining words\footnote{Buffer words are often described as ``unread'' words when describing the algorithm, but for incrementality purposes we need to count them as ``accessed'' if they are used as features, as the parser implementation is using them for prediction.} which always takes the form $\beta_i = w_{i} \ldots w_{|w|}$ for some $i$, and $A$ is the partial parse at that state. This parser is strongly incremental, as the way in which the algorithm constructs dependencies (in a strictly left-to-right manner) means that a configuration with buffer $\beta_i$ can hold every possible partial parse for the prefix $w_1 \ldots w_{i}$. The parser's delay depends on the number of buffer words used as lookahead features in the implementation. In our case, this is only one (we only use the first stack word and the first buffer word) so we can obtain partial parses for $w_1 \ldots w_{i}$ accessing only $w_1 \ldots w_{i}$, hence the delay is 0. Equivalently to sequence-labeling decoders, for models with delay $k>0$, we access a concatenated vector of the form $w_i \cdot ... \cdot w_{i+k-1} \cdot w_{i+k}$. For prediction of transitions, we again use a 1-layered feed-forward network.

\section{Experiments}

We choose 12 diverse treebanks from UD 2.11 \cite{nivre-etal-2020-universal}, supported by the tested LLMs. We test all possible combinations of encoders and decoders. As a well-known baseline, we use the biaffine \citep[DM;][]{Biaffine} parser in \texttt{supar}. We use unlabelled attachment score (UAS) for evaluation. Labelled (LAS) results and individual treebank results are in Appendix \ref{sec:appendix}.

Table~\ref{table-uas-results} shows an aggregated summary of the results with strict delay zero. It shows that fully incremental models considerably lag behind their counterparts with bidirectional (control) encoders: the best fully incremental model for each language is 11.2 UAS points behind on average (sd: 5.0) with respect to the corresponding best control model. There is large inter-linguistic variability, Telugu (te\textsubscript{MTG}) being especially amenable to incremental processing, 5.3 UAS points behind, and the opposite for Chinese (zh\textsubscript{GSD}), 23.1 points behind. Our incremental-decoder-only models with LLMs as encoders are competitive against the BiLSTM-based version of the baseline (BiLSTM encoder, biaffine decoder), surpassing it on 7 out of 12 languages. However, they are a few points behind with respect to a version of the biaffine parser using RoBERTa encodings (which can be taken as a state-of-the-art system), consistent with existing comparisons of sequence-labeling parsers and biaffine parsers~\citep{anderson-gomez-rodriguez-2021-modest}. Put together, this seems to suggest that the challenge of incrementality falls mostly on the encoding side. If we focus on comparing different strongly incremental models we see that, as expected, forward-looking encodings suffer greatly from incremental encoders.

\begin{table}[tpb!]
    \centering
    \scriptsize
    \addtolength{\tabcolsep}{-2.2pt}
    \def\arraystretch{1.3}
    \label{average-uas}
    \input{tables/aacl-table-uas}
    \caption{\label{table-uas-results} UAS scores paired with best forward looking (\textbf{fl}) and non-forward looking (\textbf{non-fl}) encodings, and transition-based (\textbf{tb}) decoder. Superscripts denote the encoder-decoder pair: LSTM (\lstm), BLOOM-560m (\bloom), mGPT (\mgpt), BiLSTM (\bilstm), XLM-RoBERTa (\xlm). Subscripts denote the best performing encoding: absolute (\abs), relative (\rel), PoS-tag-based (\relpos), 1-planar (\planari) and 2-planar (\planarii). Macro-average ($\mu$) and baseline performance (\textbf{DM}, for Dozat and Manning) with different encoders (\bilstm, \xlm) are included. Language abbreviations come from ISO 639-1 (Table~\ref{branching-percent} in the Appendix).}
\end{table}

Table \ref{table-uas-results-delay} compares the results from Table \ref{table-uas-results} against the corresponding models using delays 1 and 2. Improvements are consistent across the board. Interestingly, moving from delay 0 to 1 already shows a clear and large increase in robustness, especially for forward-looking encodings: the average gap between these and non-forward-looking encodings goes from over 10 points with delay 0 to nonexistent with delay 1, although considerable gaps remain in some languages like Chinese (zh\textsubscript{GSD}) or English (en\textsubscript{EWT}).

\begin{table}[tbp!]
    \centering
    \scriptsize
    \def\arraystretch{1.3}
    \input{tables/aacl-table-uas-delay}
    \caption{\label{table-uas-results-delay} UAS scores with delay 1 and 2. Notation as in Table \ref{table-uas-results}. Subscripts denote performance boost over zero-delay fully incremental results from Table \ref{table-uas-results}.} 
\end{table}

\begin{figure}[tbp]
    \hspace{-1em}
    \includegraphics[width=0.5\textwidth]{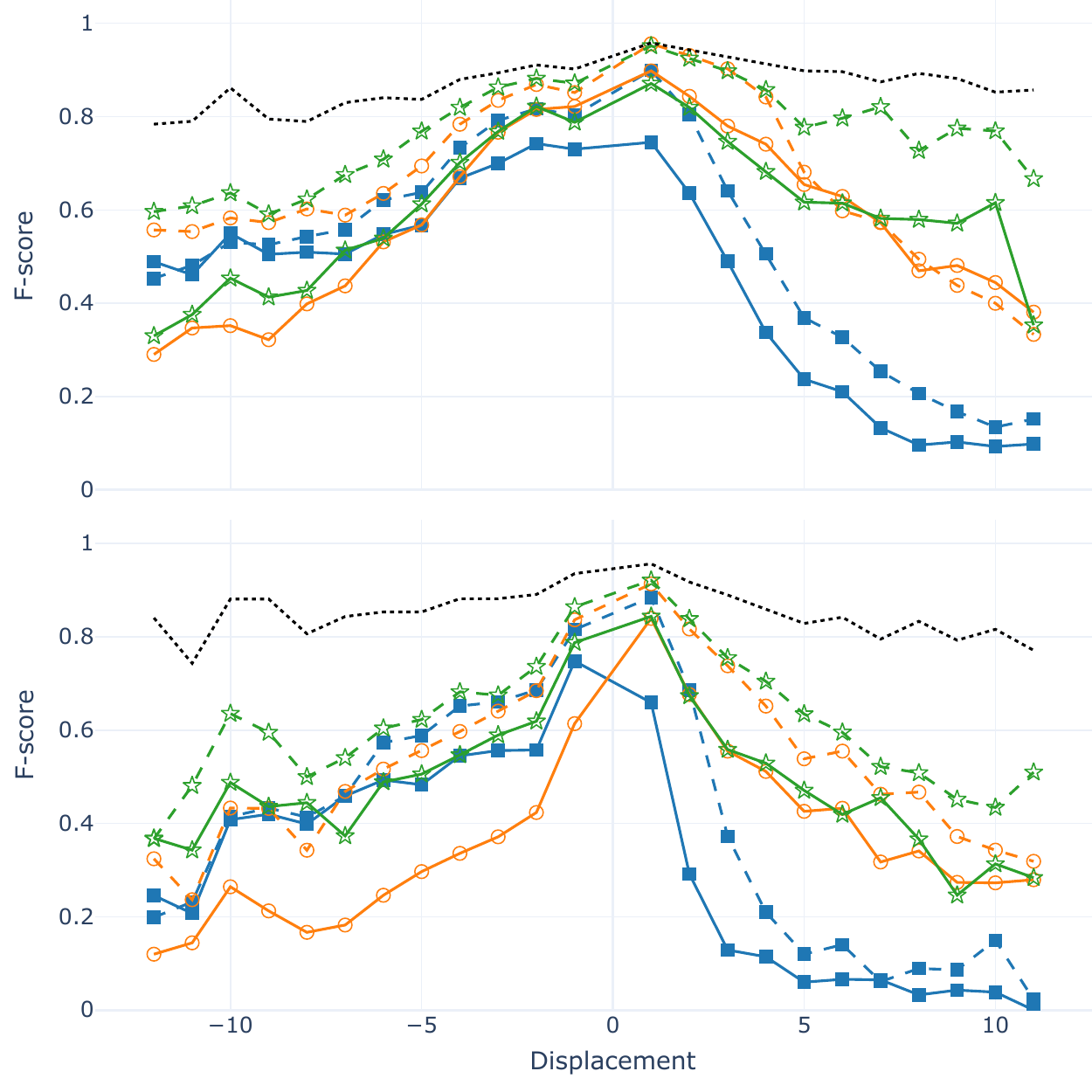}
    \caption{\label{fig:displacement} Displacement performance  (English above, Chinese below) 
    for the (fully incremental) models specified in Table \ref{table-uas-results} with delay zero (solid lines) and two (dashed lines). Different symbols and colors denote forward-looking (\fl), non-forward looking (\nonfl) and transition-based (\tb) decoders. DM\SP{\xlm} performance is included with gray dotted lines.}
\end{figure}

Finally, Figure \ref{fig:displacement} complements Table \ref{table-uas-results-delay} with an analysis of the F-score with respect to dependency displacement (signed distance) for English and Chinese, chosen because they yielded the largest improvements when using positive delay. In particular, the figure shows that the lower performance of delay zero models is mainly due to poor performance of forward-looking encodings on leftward dependencies (right half of figure), and that a small positive delay already translates into clear improvements, even for long-distance dependencies.

\section{Conclusion}

We evaluated modern neural NLP architectures for incremental dependency parsing across multiple languages, using various encoders and decoders. We have found that said architectures are not adequate to model incrementality, at least in the absence of specific adaptations. Strongly incremental models with no delay yield accuracies about 10 points below competitive non-incremental baselines. While this gap narrows when adding a 2-word lookahead, it is still about 5 points, contrasting with the situation in pre-deep-learning times, when incremental parsers  were competitive (cf. \cite{zhang-nivre-2011-transition}). 
The results suggest that much of the accuracy improvements in parsing obtained in recent years hinge on bidirectionality, deviating from human processing.

Accurate incremental parsing should in theory be possible (as the human example shows). Incremental processing is useful both for practical applications~\citep{Koehn19}, specially those involving real-time speech~\citep{coman,ekstedt-skantze-2021-projection}; as well as for cognitive modeling~\citep{demberg2019cognitive,stanojevic-etal-2021-modeling}. Thus, we believe that designing architectures that work well in a strongly incremental setting is an important open challenge in NLP. In this respect, techniques like using tentative predictions of future words made by autoregressive language models as a substitute for delay~\citep{madureira-schlangen-2020-incremental} might be helpful. It is also conceivable that accuracy losses might not be solvable by better unidirectional scoring systems, and thus alternatives such as better search or methods that revise earlier decisions are also worth exploring.

\section*{Limitations}

\paragraph{Limited physical resources} We have no access to large computing infrastructures or a budget to scale services in the cloud. We had access to a few internal servers, for a total of 6 NVIDIA GeForce RTX 3090 (24GB each), and temporally we also obtained access to a NVIDIA A100 GPU (80GB), as well as a workstation for quick debugging. This restricts the number and size of models that we can try. In particular, we could train in reasonable amounts of time the smallest BLOOM language model (560M parameters). It was possible for us to fit up to the 3B version on the A100 GPU with a minimal batch size, but the amount of time that it took to train a model made it unfeasible to carry out a multilingual study like the one proposed in this work. Still, preliminary results showed that these larger BLOOM models were not contributing to significantly improve the performance. In this respect, we know that scaling \emph{a lot} can play an important role, and that the standard BLOOM model is the 176B version. However, a model of that size is completely out of our economic and computing resources. Yet, we feel our study with smaller models is equally, or even more relevant, since it represents effectively the resources at hand for most companies and academic institutions.

\paragraph{Delay parameter}

Incremental parsers have a \emph{delay} parameter that models how far beyond word $i$ the parser can access to generate a partial parse for $w_1 \ldots w_i$. For our main experiment we set the delay to $0$, although we also provide results with delay $1$ and $2$. If we aim for psycholinguistic plausibility, there cannot be a single one-size-fits-all value for the delay, as the time taken by humans to parse linguistic input can be influenced by various factors like language, word length, reading/speaking speed, language proficiency, etc.; so any choice of delay is necessarily a simplification. However, evidence seems to point to human parsing generally being very fast, with latencies in the range of 100-250 ms (see for example \citet{Pulvermuller09,Bemis2011}). Hence our choice of delay 0 as the safest option, and we also present experiments with $1$ and $2$ to show what happens when a small lag between the input and the parse is accounted for.

\paragraph{Scope of definition}

Our definition of strong incrementality only applies to monotonic parsers. This is a deliberate choice: if we allowed non-monotonicity (i.e., removing or modifying dependencies from previous partial parses), then the definition would allow for a hypothetical parser that removes all partial output upon reading the last word and replaces it with a brand new parse generated with access to the whole sentence, which would be incremental in name only and render any comparison between incremental and non-incremental parsers moot.

While there might be alternative ways to restrict the definition to avoid this problem (e.g. restrict each step to be $O(1)$), these would come with their own limitations (e.g., excluding neural architectures where obtaining each word's vector representation is $O(n)$, or transition-based parsers with quadratic complexities). Thus, we believe that our definition is a good compromise for our purposes, as it is simple, unambiguous and implementation-independent within the realm of monotonic parsing.

Comparing non-monotonic parsers is a different undertaking as it not only would require a different definition of incrementality, but also evaluation metrics focused on partial parse accuracy rather than final LAS/UAS. But that is orthogonal to comparing incremental to non-incremental parsers (as partial parse accuracy is not even well-defined for some non-incremental parsers that do not have intermediate states) and lies outside the scope of this paper.

\paragraph{Differences in incremental processing between humans and machines} Currently, despite research efforts, a comprehensive understanding of why humans excel at incremental processing compared to machines remains elusive. This issue also constrains our options for analysis. In this regard, the proficiency of humans at incremental language processing likely stems from adaptation in the context of cognitive constraints, having to understand real-time input with limited working memory which forces eager processing (see e.g.~\citealt{christiansen_chater_2016}). From a different perspective, \citet{wilcox-etal-2021-targeted} showed that both humans and models exhibit increased processing difficulty in ungrammatical sentences. However, language models consistently underestimate the magnitude of this difficulty compared to humans, particularly in predicting longer reaction times for syntactic violations.

\section*{Acknowledgments}

We acknowledge the European Research Council (ERC), which has funded this research under the Horizon Europe research and innovation programme (SALSA, grant agreement No 101100615), ERDF/MICINN-AEI (SCANNER-UDC, PID2020-113230RB-C21), Xunta de Galicia (ED431C 2020/11), Cátedra CICAS (Sngular, University of A Coruña), and Centro de Investigación de Galicia ‘‘CITIC’’, funded by the Xunta de Galicia through the collaboration agreement between the Consellería de Cultura, Educación, Formación Profesional e Universidades and the Galician universities for the reinforcement of the research centres of the Galician University System (CIGUS).

\bibliography{anthology,custom}
\bibliographystyle{acl_natbib}

\appendix

\section{Appendix}
\label{sec:appendix}

\subsection{Additional results}
Tables \ref{table-las-results} and \ref{table-las-results-delay} show the aggregate results according to LAS.
Tables \ref{arabic-results} to \ref{vietnamese-results} illustrate the performance of every individual encoder-decoder pair in each treebank test set.

\begin{table}[thbp!]
    \centering
    \scriptsize
    \addtolength{\tabcolsep}{-2.2pt}
    \def\arraystretch{1.3}
    \input{tables/aacl-table-las}

    \caption{\label{table-las-results} LAS scores. Notation comes from Table \ref{table-uas-results}.} 
\end{table}

\begin{table}[tbp!]
    \centering
    \scriptsize
    \def\arraystretch{1.3}
    \input{tables/aacl-table-las-delay}
    \caption{\label{table-las-results-delay} LAS scores with delay one and two. Notation comes from Table \ref{table-uas-results-delay}.} 
\end{table}

\input{results/arabic-padt/complete-delay}
\input{results/basque-bdt/complete-delay}
\input{results/chinese-gsd/complete-delay}
\input{results/english-ewt/complete-delay}
\input{results/french-gsd/complete-delay}
\input{results/hindi-hdtb/complete-delay}
\input{results/indonesian-gsd/complete-delay}
\input{results/marathi-ufal/complete-delay}
\input{results/spanish-ancora/complete-delay}
\input{results/tamil-ttb/complete-delay}
\input{results/telugu-mtg/complete-results}
\input{results/vietnamese-vtb/complete-delay}

\subsection{Training hyperparameters and model configuration}

Table \ref{hyper-arch} shows the hyperparameters selected in the training process of each encoder. For LSTMs and BiLSTMs, words and PoS tags were mapped to a 300-dimensional and 100-dimensional vector representation, respectively. 
 Word information at the character level was represented with the last hidden state of dimension 100 from a charLSTM. These three representations were concatenated and projected to the encoder hidden size. For pretrained encoders (BLOOM, mGPT and XLM-RoBERTa), we get their last layer representations. Then, they are linearly projected to a smaller dimensionality of size 100.

 Table \ref{hyper-train} shows the training configuration per architecture: LSTM and BiLSTM weights were fitted with Adam optimizer ($\beta_0=0.9$, $\beta_1=0.9$, $\varepsilon=1e-12$) and pretrained encoders with AdamW ($\beta_0=0.9$, $\beta_1=0.9$, $\varepsilon=1e-12$). Batch size was adapted by the number of parameters of the encoders and the size of the treebank: in small encoders (LSTMs and BiLSTMs) data was distributed in batches of size 600, while pretrained encoders were trained with batches of size 2000.

\begin{table}[thbp!]
    \centering
    \scriptsize
    \addtolength{\tabcolsep}{-2.5pt}
    \def\arraystretch{1.4}
    \begin{tabular}{c|ccc|cc}
    \multirow{2}{*}{\bf Hyperparameter} & \multicolumn{3}{c|}{\bf Fully incremental} & \multicolumn{2}{c}{\bf Non incremental}  \\
    & LSTM & BLOOM & mGPT & BiLSTM & XLM  \\ 
    \hline
    Word emb size & 300 & 1 & 1 & 300 & 1  \\
    PoS-feats emb size & 100 & x & x& 100 & x \\
    Character emb size & 50 & x & x & 50 & x\\ 
    CharLSTM hidden & 100 & x & x & 100 & x\\
    Num. layers & 4 & 1 & 1 & 4 & 1 \\
    Encoder hidden & 400 & 100 & 100 & 400 & 100\\
\end{tabular}
    \caption{\label{hyper-arch}Architecture design choices for different encoders} 
\end{table}

\begin{table}[thbp!]
    \centering
    \scriptsize
    \addtolength{\tabcolsep}{-2.5pt}
    \def\arraystretch{1.4}
    \begin{tabular}{c|ccc|cc}
    \multirow{2}{*}{\bf Hyperparameter} & \multicolumn{3}{c|}{\bf Fully incremental} & \multicolumn{2}{c}{\bf Non incremental}  \\
    & LSTM & BLOOM & mGPT & BiLSTM & XLM  \\ 
    \hline
    lr & 1e-3 & 5e-5 & 5e-5 & 1e-3 & 5e-5\\
    optimizer & Adam & AdamW & AdamW & Adam & AdamW\\
    decay type & Exponential & Linear & Linear & Exponential & Linear\\
    decay value & 0.1 & 0.5 & 0.1 & 0.1 & 0.5 \\
    epochs & 200 & 30 & 30 & 200 & 30\\ 
    batch size & \textasciitilde 6000 &  \textasciitilde 2000 & \textasciitilde 500 & \textasciitilde 6000 & \textasciitilde 2000\\
\end{tabular}
    \caption{\label{hyper-train}Training hyper-parameters for different encoders} 
\end{table}

\subsection{Other statistics about the treebanks}

Table \ref{branching-percent} shows treebank language abbreviations and the percentage of arcs that point to the left and to the right in each treebank.

\begin{table}[thbp!]
    \centering
    \scriptsize
    \addtolength{\tabcolsep}{-2.5pt}
    \def\arraystretch{1.4}
    \begin{tabular}{c|ccc}
    \hline
    \textbf{ISO code} &  \textbf{Language}    &   \% \textbf{left-arcs}    & \% \textbf{right-arcs} \\
    \hline

    ar\textsubscript{PADT}  & Arabic        &   30.46\% & 69.54\% \\
    eu\textsubscript{BDT}   & Basque        &   49.22\% & 50.78\% \\
    zh\textsubscript{GSD}   & Chinese       &   63.67\% & 36.33\% \\
    en\textsubscript{EWT}   & English       &   57.18\% & 42.82\% \\
    fr\textsubscript{GSD}   & French        &   54.72\% & 45.28\% \\
    hi\textsubscript{HDTB}  & Hindi         &    55.6\% &  44.4\% \\
    in\textsubscript{GSD}   & Indonesian    &   37.75\% & 62.25\% \\
    mr\textsubscript{UFAL}  & Marathi       &   51.34\% & 48.66\% \\
    es\textsubscript{ANC}   & Spanish       &   54.43\% & 45.57\% \\
    ta\textsubscript{TTB}   & Tamil         &   68.56\% & 31.44\% \\
    te\textsubscript{MTG}   & Telugu        &   54.28\% & 45.72\% \\
    vi\textsubscript{VTB}   & Vietnamese    &   40.99\% & 59.01\% \\
    \hline
    \end{tabular}
    \caption{\label{branching-percent} Statistics of treebanks retrieved in our multilingual benchmark.} 
\end{table}

\end{document}

%% file: tables/aacl-table-uas.tex
\begin{tabular}{p{0.57cm}|p{0.65cm}p{0.65cm}p{0.5cm}|p{0.7cm}p{0.7cm}p{0.5cm}|cc|}
    \multirow{3}{*}{}&\multicolumn{3}{c|}{\bf Fully incremental}&\multicolumn{3}{c|}{\bf Incremental decoder} & \multicolumn{2}{c|}{\bf DM}\\
    & \multicolumn{1}{c}{\bf fl} & \multicolumn{1}{c}{\bf non-fl} &\multicolumn{1}{c|}{\bf tb}& \multicolumn{1}{c}{\bf fl} & \multicolumn{1}{c}{\bf non-fl} & \multicolumn{1}{c|}{\bf tb}& \bilstm & \multicolumn{1}{c|}{\xlm}\\
    \hline
    ar\textsubscript{PADT} &  75.7\SPSB{\lstm}{\rel} &       76.8\SPSB{\mgpt}{\planari} &     79.6\SP{\bloom} &     84.6\SPSB{\xlm}{\rel} &     88.0\SPSB{\xlm}{\planarii} &     86.9\SP{\xlm}  &   86.9     &    91.0 \\
    eu\textsubscript{BDT}  &  62.0\SPSB{\mgpt}{\pos} &       73.0\SPSB{\mgpt}{\planari} &      71.0\SP{\mgpt} &  87.0\SPSB{\bilstm}{\rel} &     87.6\SPSB{\xlm}{\planarii} &     86.6\SP{\xlm}  &   88.2     &    88.6 \\
    zh\textsubscript{GSD}  &  51.1\SPSB{\lstm}{\rel} &     64.4\SPSB{\bloom}{\planarii} &      64.1\SP{\mgpt} &     83.4\SPSB{\xlm}{\rel} &      87.5\SPSB{\xlm}{\planari} &     85.3\SP{\xlm}  &   86.7     &    90.7 \\
    en\textsubscript{EWT}  &  61.5\SPSB{\lstm}{\pos} &      74.7\SPSB{\mgpt}{\planarii} &      72.9\SP{\mgpt} &     90.1\SPSB{\xlm}{\rel} &      91.6\SPSB{\xlm}{\planari} &     89.5\SP{\xlm}  &   90.2     &    92.7 \\
    fr\textsubscript{GSD}  &  70.9\SPSB{\lstm}{\rel} &       84.4\SPSB{\mgpt}{\planari} &      84.7\SP{\lstm} &     92.3\SPSB{\xlm}{\abs} &     94.7\SPSB{\xlm}{\planarii} &     91.6\SP{\xlm}  &   93.5     &    95.0 \\
    hi\textsubscript{HDTB} &  67.1\SPSB{\lstm}{\pos} &       83.5\SPSB{\mgpt}{\planari} &      83.2\SP{\lstm} &     94.3\SPSB{\xlm}{\abs} &     95.3\SPSB{\xlm}{\planarii} &     93.8\SP{\xlm}  &   95.5     &    95.7 \\
    id\textsubscript{GSD}  &  73.0\SPSB{\lstm}{\rel} &       77.9\SPSB{\lstm}{\planari} &      78.6\SP{\lstm} &  84.5\SPSB{\bilstm}{\rel} &     86.3\SPSB{\xlm}{\planarii} &     85.1\SP{\xlm}  &   88.5     &    89.6 \\
    mr\textsubscript{UFAL} &  64.1\SPSB{\lstm}{\pos} &      69.7\SPSB{\lstm}{\planarii} &      65.8\SP{\lstm} &  75.7\SPSB{\bilstm}{\rel} &   75.2\SPSB{\bilstm}{\planari} &  76.0\SP{\bilstm}  &   79.2     &    81.7 \\
    es\textsubscript{ANC}  &  67.9\SPSB{\lstm}{\rel} &      83.2\SPSB{\mgpt}{\planarii} &      82.9\SP{\mgpt} &     92.0\SPSB{\xlm}{\abs} &      93.4\SPSB{\xlm}{\planari} &     91.2\SP{\xlm}  &   93.1     &    94.3 \\
    ta\textsubscript{TTB}  &  59.1\SPSB{\lstm}{\pos} &       67.3\SPSB{\lstm}{\planari} &      69.7\SP{\lstm} &  71.4\SPSB{\bilstm}{\rel} &      75.8\SPSB{\xlm}{\planari} &     78.6\SP{\xlm}  &   77.2     &    80.0 \\
    te\textsubscript{MTG}  &  73.8\SPSB{\lstm}{\pos} &      85.0\SPSB{\lstm}{\planarii} &      80.4\SP{\lstm} &  89.9\SPSB{\bilstm}{\rel} &  90.3\SPSB{\bilstm}{\planarii} &  89.0\SP{\bilstm}  &   89.2     &    94.5 \\
    vi\textsubscript{VTB}  &  57.3\SPSB{\lstm}{\rel} &      64.5\SPSB{\lstm}{\planarii} &      64.5\SP{\lstm} &  72.7\SPSB{\bilstm}{\rel} &      74.1\SPSB{\xlm}{\planari} &     76.0\SP{\xlm}  &   77.0     &    80.2 \\
    \hline 
    $\mu$                    &                    65.3 &                             75.4 &                74.8 &                      84.8 &                           84.6 &              85.8  &   87.1     &    89.5 
\end{tabular}

%% file: tables/aacl-table-uas-delay.tex
\begin{tabular}{p{0.7cm}|p{0.6cm}p{0.7cm}p{0.63cm}|p{0.6cm}p{0.7cm}p{0.63cm}|}
    \multirow{3}{*}{} & \multicolumn{3}{c|}{\bf Fully incremental delay 1} & \multicolumn{3}{c|}{\bf Fully incremental delay 2}\\
    & \multicolumn{1}{c}{\bf fl} & \multicolumn{1}{c}{\bf non-fl} &\multicolumn{1}{c|}{\bf tb}& \multicolumn{1}{c}{\bf fl} & \multicolumn{1}{c}{\bf non-fl} & \multicolumn{1}{c|}{\bf tb} \\
    \hline
    ar\textsubscript{PADT} &   84.2\SPSB{\mgpt\pos}{\textcolor{green}{+8.5}} &  80.2\SPSB{\mgpt\planarii}{\textcolor{green}{+3.4}} &   79.9\SPSB{\mgpt}{\textcolor{green}{+0.3}} &   83.9\SPSB{\bloom\pos}{\textcolor{green}{+8.2}} &    82.9\SPSB{\mgpt\planari}{\textcolor{green}{+6.1}} &   80.0\SPSB{\mgpt}{\textcolor{green}{+0.4}} \\
    eu\textsubscript{BDT}  &  78.4\SPSB{\mgpt\pos}{\textcolor{green}{+16.4}} &   76.9\SPSB{\mgpt\planari}{\textcolor{green}{+3.9}} &   75.7\SPSB{\mgpt}{\textcolor{green}{+4.7}} &   80.0\SPSB{\mgpt\pos}{\textcolor{green}{+18.0}} &    80.1\SPSB{\mgpt\planari}{\textcolor{green}{+7.1}} &   76.9\SPSB{\mgpt}{\textcolor{green}{+5.9}} \\
    zh\textsubscript{GSD}  &  64.3\SPSB{\mgpt\pos}{\textcolor{green}{+13.2}} &  73.5\SPSB{\mgpt\planarii}{\textcolor{green}{+9.1}} &   72.8\SPSB{\mgpt}{\textcolor{green}{+8.7}} &  68.4\SPSB{\bloom\pos}{\textcolor{green}{+17.3}} &  74.8\SPSB{\mgpt\planarii}{\textcolor{green}{+10.4}} &  75.6\SPSB{\mgpt}{\textcolor{green}{+11.5}} \\
    en\textsubscript{EWT}  &  81.9\SPSB{\mgpt\pos}{\textcolor{green}{+20.4}} &  88.1\SPSB{\mgpt\planari}{\textcolor{green}{+13.4}} &  83.3\SPSB{\mgpt}{\textcolor{green}{+10.4}} &   85.6\SPSB{\mgpt\pos}{\textcolor{green}{+24.1}} &   88.7\SPSB{\mgpt\planari}{\textcolor{green}{+14.0}} &  85.2\SPSB{\mgpt}{\textcolor{green}{+12.3}} \\
    fr\textsubscript{GSD}  &  84.4\SPSB{\mgpt\pos}{\textcolor{green}{+13.5}} &  86.2\SPSB{\mgpt\planarii}{\textcolor{green}{+1.8}} &   87.5\SPSB{\mgpt}{\textcolor{green}{+2.8}} &   87.6\SPSB{\mgpt\pos}{\textcolor{green}{+16.7}} &    89.2\SPSB{\mgpt\planari}{\textcolor{green}{+4.8}} &  86.7\SPSB{\bloom}{\textcolor{green}{+2.0}} \\
    hi\textsubscript{HDTB} &  82.5\SPSB{\mgpt\pos}{\textcolor{green}{+15.4}} &   86.2\SPSB{\mgpt\planari}{\textcolor{green}{+2.7}} &   88.7\SPSB{\mgpt}{\textcolor{green}{+5.5}} &   85.8\SPSB{\mgpt\pos}{\textcolor{green}{+18.7}} &    90.1\SPSB{\mgpt\planari}{\textcolor{green}{+6.6}} &   88.9\SPSB{\lstm}{\textcolor{green}{+5.7}} \\
    id\textsubscript{GSD}  &   80.8\SPSB{\mgpt\pos}{\textcolor{green}{+7.8}} &   80.4\SPSB{\mgpt\planari}{\textcolor{green}{+2.5}} &   79.2\SPSB{\lstm}{\textcolor{green}{+0.6}} &    81.9\SPSB{\mgpt\pos}{\textcolor{green}{+8.9}} &   82.1\SPSB{\mgpt\planarii}{\textcolor{green}{+4.2}} &   79.6\SPSB{\mgpt}{\textcolor{green}{+1.0}} \\
    mr\textsubscript{UFAL} &  73.5\SPSB{\bloom\pos}{\textcolor{green}{+9.4}} &     65.8\SPSB{\mgpt\planari}{\textcolor{red}{-3.9}} &   72.7\SPSB{\mgpt}{\textcolor{green}{+6.9}} &   72.3\SPSB{\bloom\pos}{\textcolor{green}{+8.2}} &    64.9\SPSB{\bloom\planarii}{\textcolor{red}{-4.8}} &   71.1\SPSB{\lstm}{\textcolor{green}{+5.3}} \\
    es\textsubscript{ANC}  &  84.9\SPSB{\mgpt\pos}{\textcolor{green}{+17.0}} &   85.4\SPSB{\mgpt\planari}{\textcolor{green}{+2.2}} &   85.2\SPSB{\mgpt}{\textcolor{green}{+2.3}} &   88.4\SPSB{\mgpt\pos}{\textcolor{green}{+20.5}} &    87.6\SPSB{\mgpt\planari}{\textcolor{green}{+4.4}} &   85.6\SPSB{\mgpt}{\textcolor{green}{+2.7}} \\
    ta\textsubscript{TTB}  &  68.5\SPSB{\bloom\pos}{\textcolor{green}{+9.4}} &    62.4\SPSB{\bloom\planari}{\textcolor{red}{-4.9}} &     67.0\SPSB{\mgpt}{\textcolor{red}{-2.7}} &   69.9\SPSB{\mgpt\pos}{\textcolor{green}{+10.8}} &     64.9\SPSB{\bloom\planari}{\textcolor{red}{-2.4}} &  71.9\SPSB{\bloom}{\textcolor{green}{+2.2}} \\
    te\textsubscript{MTG}  &  85.0\SPSB{\mgpt\pos}{\textcolor{green}{+11.2}} &     85.0\SPSB{\mgpt\planarii}{\textcolor{red}{0.0}} &   87.4\SPSB{\mgpt}{\textcolor{green}{+7.0}} &   86.7\SPSB{\mgpt\pos}{\textcolor{green}{+12.9}} &    89.6\SPSB{\mgpt\planari}{\textcolor{green}{+4.6}} &   90.0\SPSB{\lstm}{\textcolor{green}{+9.6}} \\
    vi\textsubscript{VTB}  &   66.6\SPSB{\mgpt\pos}{\textcolor{green}{+9.3}} &    63.8\SPSB{\mgpt\planarii}{\textcolor{red}{-0.7}} &    64.0\SPSB{\bloom}{\textcolor{red}{-0.5}} &   68.3\SPSB{\mgpt\pos}{\textcolor{green}{+11.0}} &   65.1\SPSB{\mgpt\planarii}{\textcolor{green}{+0.6}} &   65.2\SPSB{\mgpt}{\textcolor{green}{+0.7}} \\ 
    \hline
    $\mu$                  &               77.9\SB{\textcolor{green}{+12.6}} &                    77.8\SB{\textcolor{green}{+2.4}} &           78.6\SB{\textcolor{green}{+3.8}}  &                79.9\SB{\textcolor{green}{+14.6}} &                     80.0\SB{\textcolor{green}{+4.6}} &            79.7\SB{\textcolor{green}{+4.9}} \\
\end{tabular}

%% file: tables/aacl-table-las.tex
\begin{tabular}{p{0.57cm}|p{0.65cm}p{0.65cm}p{0.5cm}|p{0.7cm}p{0.7cm}p{0.5cm}|cc|}
    \multirow{3}{*}{}&\multicolumn{3}{c|}{\bf Fully incremental}&\multicolumn{3}{c|}{\bf Incremental decoder} & \multicolumn{2}{c|}{\bf DM}\\
    & \multicolumn{1}{c}{\bf fl} & \multicolumn{1}{c}{\bf non-fl} &\multicolumn{1}{c|}{\bf tb}& \multicolumn{1}{c}{\bf fl} & \multicolumn{1}{c}{\bf non-fl} & \multicolumn{1}{c|}{\bf tb}& \bilstm & \multicolumn{1}{c|}{\xlm} \\
    \hline
    ar\textsubscript{PADT} &   70.1\SPSB{\lstm}{\rel} &   70.5\SPSB{\mgpt}{\planari} &  69.5\SP{\mgpt} &     79.0\SPSB{\xlm}{\rel} &     82.3\SPSB{\xlm}{\planarii} &     78.9\SP{\xlm} &     81.4 &                      85.9 \\
    eu\textsubscript{BDT}  &   56.4\SPSB{\mgpt}{\pos} &   63.0\SPSB{\mgpt}{\planari} &  60.4\SP{\mgpt} &  82.1\SPSB{\bilstm}{\rel} &     84.4\SPSB{\xlm}{\planarii} &     76.0\SP{\xlm} &     82.4 &                      85.0 \\
    zh\textsubscript{GSD}  &   43.9\SPSB{\lstm}{\rel} &  51.4\SPSB{\mgpt}{\planarii} &  54.6\SP{\mgpt} &     80.5\SPSB{\xlm}{\rel} &      84.1\SPSB{\xlm}{\planari} &     77.5\SP{\xlm} &     83.9 &                      87.6 \\
    en\textsubscript{EWT}  &   59.0\SPSB{\lstm}{\pos} &   67.5\SPSB{\lstm}{\planari} &  65.4\SP{\mgpt} &     87.9\SPSB{\xlm}{\rel} &      89.1\SPSB{\xlm}{\planari} &     84.5\SP{\xlm} &     88.0 &                      90.1 \\
    fr\textsubscript{GSD}  &   67.1\SPSB{\lstm}{\rel} &  78.2\SPSB{\mgpt}{\planarii} &  78.1\SP{\mgpt} &     90.0\SPSB{\xlm}{\abs} &     92.4\SPSB{\xlm}{\planarii} &     87.0\SP{\xlm} &     90.0 &                      92.1 \\
    hi\textsubscript{HDTB} &   60.7\SPSB{\lstm}{\pos} &  69.6\SPSB{\mgpt}{\planarii} &  73.1\SP{\mgpt} &     91.3\SPSB{\xlm}{\abs} &     92.3\SPSB{\xlm}{\planarii} &     86.0\SP{\xlm} &     92.6 &                      92.4 \\
    id\textsubscript{GSD}  &   68.2\SPSB{\lstm}{\rel} &   72.1\SPSB{\lstm}{\planari} &  67.7\SP{\lstm} &  79.5\SPSB{\bilstm}{\rel} &     79.2\SPSB{\xlm}{\planarii} &  73.5\SP{\bilstm} &     83.2 &                      82.5 \\
    mr\textsubscript{UFAL} &  48.3\SPSB{\bloom}{\pos} &   45.9\SPSB{\mgpt}{\planari} &  36.4\SP{\mgpt} &     64.6\SPSB{\xlm}{\pos} &      62.6\SPSB{\xlm}{\planari} &     61.6\SP{\xlm} &     59.6 &                      72.8 \\
    es\textsubscript{ANC}  &   63.8\SPSB{\lstm}{\rel} &  78.9\SPSB{\mgpt}{\planarii} &  76.3\SP{\mgpt} &     90.0\SPSB{\xlm}{\abs} &     91.6\SPSB{\xlm}{\planarii} &     85.5\SP{\xlm} &     89.6 &                      92.3 \\
    ta\textsubscript{TTB}  &   50.6\SPSB{\lstm}{\pos} &   52.6\SPSB{\lstm}{\planari} &  42.7\SP{\mgpt} &  62.3\SPSB{\bilstm}{\rel} &  66.3\SPSB{\bilstm}{\planarii} &     64.8\SP{\xlm} &     66.2 &                      68.9 \\
    te\textsubscript{MTG}  &   65.6\SPSB{\mgpt}{\pos} &   67.3\SPSB{\mgpt}{\planari} &  53.5\SP{\lstm} &     79.8\SPSB{\xlm}{\rel} &      81.7\SPSB{\xlm}{\planari} &  64.6\SP{\bilstm} &     69.4 &                      87.5 \\
    vi\textsubscript{VTB}  &   45.5\SPSB{\lstm}{\rel} &  50.8\SPSB{\lstm}{\planarii} &  43.8\SP{\lstm} &  60.3\SPSB{\bilstm}{\rel} &     61.5\SPSB{\xlm}{\planarii} &     56.8\SP{\xlm} &     62.9 &                      67.3 \\
    \hline 
    $\mu$                  &                     58.3 &                         64.0 &            60.1 &                      78.9 &                           80.6 &              74.7 &      79.1 &                     83.7     
\end{tabular}

%% file: tables/aacl-table-las-delay.tex
\begin{tabular}{p{0.7cm}|p{0.6cm}p{0.7cm}p{0.63cm}|p{0.6cm}p{0.7cm}p{0.63cm}|}
    \multirow{3}{*}{} & \multicolumn{3}{c|}{\bf Fully incremental delay 1} & \multicolumn{3}{c|}{\bf Fully incremental delay 2}\\
    & \multicolumn{1}{c}{\bf fl} & \multicolumn{1}{c}{\bf non-fl} &\multicolumn{1}{c|}{\bf tb}& \multicolumn{1}{c}{\bf fl} & \multicolumn{1}{c}{\bf non-fl} & \multicolumn{1}{c|}{\bf tb} \\
    \hline
    ar\textsubscript{PADT} &    78.8\SPSB{\mgpt\pos}{\textcolor{green}{+8.7}} &   74.6\SPSB{\mgpt\planarii}{\textcolor{green}{+4.1}} &     68.1\SPSB{\mgpt}{\textcolor{red}{-1.4}} &    78.3\SPSB{\mgpt\pos}{\textcolor{green}{+8.2}} &    77.1\SPSB{\mgpt\planari}{\textcolor{green}{+6.6}} &     68.7\SPSB{\mgpt}{\textcolor{red}{-0.8}} \\
    eu\textsubscript{BDT}  &   73.5\SPSB{\mgpt\pos}{\textcolor{green}{+17.1}} &    71.2\SPSB{\mgpt\planari}{\textcolor{green}{+8.2}} &     59.4\SPSB{\mgpt}{\textcolor{red}{-1.0}} &   75.2\SPSB{\mgpt\pos}{\textcolor{green}{+18.8}} &   75.2\SPSB{\mgpt\planari}{\textcolor{green}{+12.2}} &   63.6\SPSB{\mgpt}{\textcolor{green}{+3.2}} \\
    zh\textsubscript{GSD}  &  60.4\SPSB{\bloom\pos}{\textcolor{green}{+16.5}} &  68.3\SPSB{\mgpt\planarii}{\textcolor{green}{+16.9}} &  61.3\SPSB{\bloom}{\textcolor{green}{+6.7}} &  64.6\SPSB{\bloom\pos}{\textcolor{green}{+20.7}} &  69.7\SPSB{\mgpt\planarii}{\textcolor{green}{+18.3}} &  75.6\SPSB{\mgpt}{\textcolor{green}{+21.0}} \\
    en\textsubscript{EWT}  &   79.1\SPSB{\mgpt\pos}{\textcolor{green}{+20.1}} &   84.2\SPSB{\mgpt\planari}{\textcolor{green}{+16.7}} &  83.3\SPSB{\mgpt}{\textcolor{green}{+17.9}} &   83.0\SPSB{\mgpt\pos}{\textcolor{green}{+24.0}} &   85.0\SPSB{\mgpt\planari}{\textcolor{green}{+17.5}} &  85.2\SPSB{\mgpt}{\textcolor{green}{+19.8}} \\
    fr\textsubscript{GSD}  &   80.9\SPSB{\mgpt\pos}{\textcolor{green}{+13.8}} &   81.9\SPSB{\mgpt\planarii}{\textcolor{green}{+3.7}} &     76.0\SPSB{\mgpt}{\textcolor{red}{-2.1}} &   84.2\SPSB{\mgpt\pos}{\textcolor{green}{+17.1}} &    85.4\SPSB{\mgpt\planari}{\textcolor{green}{+7.2}} &     75.9\SPSB{\mgpt}{\textcolor{red}{-2.2}} \\
    hi\textsubscript{HDTB} &   78.1\SPSB{\mgpt\pos}{\textcolor{green}{+17.4}} &   79.6\SPSB{\mgpt\planari}{\textcolor{green}{+10.0}} &   77.4\SPSB{\mgpt}{\textcolor{green}{+4.3}} &   82.0\SPSB{\mgpt\pos}{\textcolor{green}{+21.3}} &   84.9\SPSB{\mgpt\planari}{\textcolor{green}{+15.3}} &   77.4\SPSB{\mgpt}{\textcolor{green}{+4.3}} \\
    id\textsubscript{GSD}  &    73.6\SPSB{\mgpt\pos}{\textcolor{green}{+5.4}} &    72.3\SPSB{\mgpt\planari}{\textcolor{green}{+0.2}} &     66.9\SPSB{\mgpt}{\textcolor{red}{-0.8}} &    75.2\SPSB{\mgpt\pos}{\textcolor{green}{+7.0}} &   74.6\SPSB{\mgpt\planarii}{\textcolor{green}{+2.5}} &   68.3\SPSB{\mgpt}{\textcolor{green}{+0.6}} \\
    mr\textsubscript{UFAL} &   58.2\SPSB{\bloom\pos}{\textcolor{green}{+9.9}} &  54.4\SPSB{\bloom\planarii}{\textcolor{green}{+8.5}} &  72.7\SPSB{\mgpt}{\textcolor{green}{+36.3}} &   57.4\SPSB{\bloom\pos}{\textcolor{green}{+9.1}} &   51.2\SPSB{\mgpt\planarii}{\textcolor{green}{+5.3}} &  66.8\SPSB{\mgpt}{\textcolor{green}{+30.4}} \\
    es\textsubscript{ANC}  &   82.4\SPSB{\mgpt\pos}{\textcolor{green}{+18.6}} &    82.7\SPSB{\mgpt\planari}{\textcolor{green}{+3.8}} &   85.2\SPSB{\mgpt}{\textcolor{green}{+8.9}} &   86.2\SPSB{\mgpt\pos}{\textcolor{green}{+22.4}} &    85.0\SPSB{\mgpt\planari}{\textcolor{green}{+6.1}} &   85.6\SPSB{\mgpt}{\textcolor{green}{+9.3}} \\
    ta\textsubscript{TTB}  &    56.4\SPSB{\mgpt\pos}{\textcolor{green}{+5.8}} &      50.3\SPSB{\mgpt\planari}{\textcolor{red}{-2.3}} &  67.0\SPSB{\mgpt}{\textcolor{green}{+24.3}} &    58.2\SPSB{\mgpt\pos}{\textcolor{green}{+7.6}} &   53.8\SPSB{\bloom\planari}{\textcolor{green}{+1.2}} &  51.8\SPSB{\bloom}{\textcolor{green}{+9.1}} \\
    te\textsubscript{MTG}  &   76.5\SPSB{\mgpt\pos}{\textcolor{green}{+10.9}} &    74.1\SPSB{\mgpt\planari}{\textcolor{green}{+6.8}} &  87.4\SPSB{\mgpt}{\textcolor{green}{+33.9}} &   76.8\SPSB{\mgpt\pos}{\textcolor{green}{+11.2}} &   79.4\SPSB{\mgpt\planari}{\textcolor{green}{+12.1}} &     48.0\SPSB{\lstm}{\textcolor{red}{-5.5}} \\
    vi\textsubscript{VTB}  &    53.6\SPSB{\mgpt\pos}{\textcolor{green}{+8.1}} &     50.2\SPSB{\mgpt\planarii}{\textcolor{red}{-0.6}} &  63.2\SPSB{\mgpt}{\textcolor{green}{+19.4}} &    54.4\SPSB{\mgpt\pos}{\textcolor{green}{+8.9}} &   51.2\SPSB{\mgpt\planarii}{\textcolor{green}{+0.4}} &   45.9\SPSB{\mgpt}{\textcolor{green}{+2.1}}  \\
    \hline
    $\mu$                  &                71.0\SB{\textcolor{green}{+12.7}} &                     70.3\SB{\textcolor{green}{+6.3}} &           72.3\SB{\textcolor{green}{+12.2}} &                73.0\SB{\textcolor{green}{+14.7}} &                     72.7\SB{\textcolor{green}{+8.7}} &            67.7\SB{\textcolor{green}{+7.6}}
\end{tabular}

%% file: results/arabic-padt/complete-delay.tex
\begin{table}[bpth]
    \centering
    \scriptsize
    \addtolength{\tabcolsep}{-3pt}
    \def\arraystretch{1.3}
    \begin{tabular}{c|c|ccc|cc|c}
        \multirow{2}{*}{\bf Metric}& \multirow{2}{*}{\bf Encoder}& \multicolumn{3}{c}{\textbf{fl}} & \multicolumn{2}{|c|}{\textbf{non-fl}} & \multirow{2}{*}{\bf TB} \\
        & &    abs-idx &    rel-idx &  PoS-idx &     1p &    2p &   \\
        \hline
        \hline
        \multirow{16}{*}{\bf UAS}
            & LSTM &  71.7 &  75.7 &  59.9 &  74.1 &  73.0 &  78.2 \\
            &     &  74.8 &  80.2 &  81.3 &  76.9 &  76.2 &  76.9 \\
            &     &  72.6 &  81.4 &  81.6 &  79.2 &  78.5 &  77.3 \\
            & mGPT &  52.6 &  74.7 &  59.5 &  76.8 &  76.4 &  78.4 \\
            &     &  65.5 &  81.7 &  84.2 &  80.0 &  80.2 &  79.9 \\
            &     &  63.8 &  82.3 &  83.6 &  82.9 &  82.4 &  80.0 \\
            & BLOOM &  43.0 &  74.4 &  58.6 &  75.6 &  75.6 &  79.6 \\
            &     &  56.5 &  81.1 &  82.8 &  79.7 &  79.3 &  79.1 \\
            &     &  55.4 &  81.5 &  83.9 &  81.9 &  82.0 &  79.9 \\
            & BiLSTM &  71.0 &  83.2 &  65.9 &  81.3 &  80.4 &  82.0 \\
            &     &  75.2 &  83.3 &  83.0 &  82.1 &  81.0 &  78.7 \\
            &     &  73.0 &  83.4 &  82.9 &  82.3 &  81.1 &  78.3 \\
            & XLM &  77.2 &  84.6 &  66.1 &  87.2 &  88.0 &  86.9 \\
            &     &  86.1 &  87.9 &  86.0 &  89.6 &  88.8 &  83.1 \\
            &     &  84.4 &  87.1 &  88.8 &  89.2 &  89.2 &  83.8 \\
        \cline{2-8}
            &\bf DM & \multicolumn{6}{c}{87.3}\\
        \hline 
        \multirow{16}{*}{\bf LAS}
            & LSTM &  67.1 &  70.1 &  55.9 &  68.8 &  67.8 &  66.3 \\
            &     &  68.6 &  73.5 &  74.9 &  71.0 &  69.6 &  59.6 \\
            &     &  66.4 &  74.7 &  75.1 &  73.0 &  72.0 &  60.7 \\
            & mGPT &  48.9 &  68.8 &  55.3 &  70.5 &  70.4 &  69.5 \\
            &     &  61.2 &  76.0 &  78.8 &  74.1 &  74.6 &  68.1 \\
            &     &  59.6 &  76.8 &  78.3 &  77.1 &  77.0 &  68.7 \\
            & BLOOM &  39.6 &  67.6 &  54.4 &  68.9 &  68.8 &  67.6 \\
            &     &  52.4 &  75.2 &  77.3 &  73.6 &  73.4 &  67.2 \\
            &     &  51.3 &  75.7 &  78.0 &  76.0 &  75.9 &  67.8 \\
            & BiLSTM &  66.8 &  77.6 &  61.9 &  76.1 &  75.2 &  73.4 \\
            &     &  69.6 &  76.6 &  76.4 &  75.8 &  74.7 &  62.8 \\
            &     &  67.6 &  76.6 &  76.3 &  76.0 &  74.6 &  61.9 \\
            & XLM &  72.5 &  79.0 &  62.7 &  81.6 &  82.3 &  78.9 \\
            &     &  80.9 &  82.3 &  81.0 &  84.5 &  83.7 &  68.6 \\
            &     &  79.6 &  82.0 &  83.9 &  84.1 &  83.8 &  71.8 \\
        \cline{2-8}
        & \bf DM & \multicolumn{6}{c}{81.8}\\
\end{tabular}
    \caption{\label{arabic-results}UAS and LAS  for the Arabic PADT treebank. Last index level corresponds to delay results. First, second and third subrow show the scores obtained with 0, 1 and 2-delay, respectively.} 
\end{table}

%% file: results/basque-bdt/complete-delay.tex
\begin{table}[bpth]
    \centering
    \scriptsize
    \addtolength{\tabcolsep}{-3pt}
    \def\arraystretch{1.3}
    \begin{tabular}{c|c|ccc|cc|c}
        \multirow{2}{*}{\bf Metric}& \multirow{2}{*}{\bf Encoder}& \multicolumn{3}{c}{\textbf{fl}} & \multicolumn{2}{|c|}{\textbf{non-fl}} & \multirow{2}{*}{\bf TB} \\
        & &    abs-idx &    rel-idx &  PoS-idx &     1p &    2p &   \\
        \hline
        \hline
        \multirow{16}{*}{\bf UAS}
            & LSTM &  59.4 &  61.7 &  61.9 &  72.0 &  71.7 &  69.7 \\
            &     &  69.0 &  70.8 &  74.0 &  70.8 &   0.0 &  73.6 \\
            &     &  68.5 &  73.9 &  75.7 &  72.9 &  74.1 &  75.5 \\
            & mGPT &  42.0 &  55.0 &  62.0 &  73.0 &  72.9 &  71.0 \\
            &     &  58.5 &  67.7 &  78.4 &  76.9 &  76.0 &  75.7 \\
            &     &  59.6 &  72.3 &  80.0 &  80.1 &  79.5 &  76.9 \\
            & BLOOM &  34.6 &  50.7 &  58.1 &  64.3 &  65.2 &  63.4 \\
            &     &  48.9 &  60.8 &  73.8 &  68.8 &  68.2 &  72.1 \\
            &     &  52.6 &  67.8 &  75.9 &  74.7 &  72.2 &  74.6 \\
            & BiLSTM &  81.5 &  87.0 &  73.3 &  83.6 &  83.7 &  83.3 \\
            &     &  72.6 &  78.5 &  78.0 &  71.3 &  71.0 &  77.5 \\
            &     &  70.7 &  79.6 &  77.3 &  71.8 &  70.6 &  76.7 \\
            & XLM &  80.9 &  83.6 &  73.7 &  87.6 &  87.6 &  86.6 \\
            &     &  84.8 &  86.4 &  86.2 &  86.4 &  85.8 &  84.4 \\
            &     &  84.9 &  85.5 &  85.2 &  86.0 &  85.6 &  85.4 \\
        \cline{2-8}
            &\bf DM & \multicolumn{6}{c}{84.8}\\
        \hline 
        \multirow{16}{*}{\bf LAS}
            & LSTM &  53.5 &  55.7 &  55.3 &  60.7 &  60.1 &  52.0 \\
            &     &  61.6 &  63.7 &  67.3 &  63.5 &   0.0 &  52.8 \\
            &     &  61.6 &  66.7 &  69.3 &  66.0 &  67.3 &  55.2 \\
            & mGPT &  38.4 &  49.6 &  56.4 &  63.0 &  62.7 &  60.4 \\
            &     &  54.4 &  63.2 &  73.5 &  71.2 &  70.4 &  59.4 \\
            &     &  55.5 &  67.8 &  75.2 &  75.2 &  74.4 &  63.6 \\
            & BLOOM &  30.8 &  44.2 &  51.4 &  53.3 &  54.2 &  50.1 \\
            &     &  43.8 &  54.8 &  66.9 &  61.9 &  60.9 &  55.5 \\
            &     &  47.5 &  62.2 &  69.7 &  68.6 &  65.8 &  58.7 \\
            & BiLSTM &  76.5 &  82.1 &  69.6 &  78.4 &  78.6 &  67.8 \\
            &     &  65.1 &  70.0 &  70.5 &  63.8 &  63.9 &  54.6 \\
            &     &  62.8 &  71.8 &  69.6 &  64.8 &  63.0 &  53.7 \\
            & XLM &  78.2 &  80.4 &  71.3 &  84.3 &  84.4 &  76.0 \\
            &     &  81.2 &  82.7 &  82.6 &  82.4 &  81.8 &  77.0 \\
            &     &  81.1 &  81.6 &  81.5 &  82.3 &  81.5 &  77.7 \\
        \cline{2-8}
        & \bf DM & \multicolumn{6}{c}{77.3}\\
\end{tabular}
    \caption{\label{basque-results}UAS and LAS  for the Basque BDT treebank. Notation comes from Table \ref{arabic-results}.} 
\end{table}

%% file: results/chinese-gsd/complete-delay.tex
\begin{table}[bpth]
    \centering
    \scriptsize
    \addtolength{\tabcolsep}{-3pt}
    \def\arraystretch{1.3}
    \begin{tabular}{c|c|ccc|cc|c}
        \multirow{2}{*}{\bf Metric}& \multirow{2}{*}{\bf Encoder}& \multicolumn{3}{c}{\textbf{fl}} & \multicolumn{2}{|c|}{\textbf{non-fl}} & \multirow{2}{*}{\bf TB} \\
        & &    abs-idx &    rel-idx &  PoS-idx &     1p &    2p &   \\
        \hline
        \hline
        \multirow{16}{*}{\bf UAS}
            & LSTM &  43.7 &  51.1 &  45.0 &  62.9 &  61.9 &  60.1 \\
            &     &  51.7 &  59.8 &  61.3 &  62.0 &  60.7 &  69.6 \\
            &     &  54.6 &  64.3 &  64.0 &  65.3 &  64.6 &  70.9 \\
            & mGPT &  33.8 &  47.6 &  44.7 &  63.4 &  63.6 &  64.1 \\
            &     &  43.3 &  57.4 &  64.3 &  72.6 &  73.5 &  72.8 \\
            &     &  40.5 &  63.0 &  67.4 &  73.6 &  74.8 &  75.6 \\
            & BLOOM &  23.7 &  45.0 &  44.4 &  63.8 &  64.4 &  60.4 \\
            &     &  32.5 &  57.3 &  64.2 &  71.5 &  71.9 &  71.4 \\
            &     &  37.9 &  62.8 &  68.4 &  72.9 &  71.7 &  73.2 \\
            & BiLSTM &  69.1 &  81.6 &  63.9 &  79.8 &  79.8 &  77.7 \\
            &     &  62.5 &  72.8 &  71.7 &  69.1 &  67.9 &  72.5 \\
            &     &  59.8 &  73.1 &  72.5 &  68.9 &  67.8 &  71.0 \\
            & XLM &  81.6 &  83.4 &  65.3 &  87.5 &  87.3 &  85.3 \\
            &     &  85.2 &  83.6 &  85.5 &  86.7 &  87.1 &  83.5 \\
            &     &  85.3 &  85.4 &  83.4 &  86.6 &  85.8 &  81.8 \\
        \cline{2-8}
            &\bf DM & \multicolumn{6}{c}{85.3}\\
        \hline 
        \multirow{16}{*}{\bf LAS}
            & LSTM &  37.5 &  43.9 &  42.0 &  50.5 &  49.4 &  50.2 \\
            &     &  46.7 &  54.0 &  56.0 &  56.5 &  54.7 &  49.8 \\
            &     &  49.5 &  58.3 &  58.6 &  59.3 &  58.5 &  52.9 \\
            & mGPT &  29.7 &  40.6 &  41.5 &  51.1 &  51.4 &  54.6 \\
            &     &  40.2 &  53.5 &  60.0 &  67.5 &  68.3 &  61.0 \\
            &     &  37.8 &  59.3 &  63.6 &  68.9 &  69.7 &  75.6 \\
            & BLOOM &  19.7 &  37.2 &  40.5 &  50.6 &  51.3 &  47.9 \\
            &     &  29.9 &  53.5 &  60.4 &  66.0 &  66.4 &  61.3 \\
            &     &  35.0 &  58.7 &  64.6 &  68.0 &  66.3 &  63.9 \\
            & BiLSTM &  66.3 &  78.5 &  61.8 &  76.7 &  76.6 &  71.4 \\
            &     &  56.4 &  66.2 &  66.4 &  64.1 &  61.9 &  55.1 \\
            &     &  54.1 &  66.6 &  67.1 &  63.4 &  62.0 &  55.4 \\
            & XLM &  78.8 &  80.5 &  63.6 &  84.1 &  84.1 &  77.5 \\
            &     &  82.3 &  80.6 &  82.8 &  83.7 &  84.3 &  74.2 \\
            &     &  82.2 &  82.2 &  81.0 &  83.3 &  82.2 &  72.4 \\
        \cline{2-8}
        & \bf DM & \multicolumn{6}{c}{82.4}\\
\end{tabular}
    \caption{\label{basque-results}UAS and LAS  for the Chinese GSD treebank. Notation comes from Table \ref{arabic-results}.} 
\end{table}

%% file: results/english-ewt/complete-delay.tex
\begin{table}[bpth]
    \centering
    \scriptsize
    \addtolength{\tabcolsep}{-3pt}
    \def\arraystretch{1.3}
    \begin{tabular}{c|c|ccc|cc|c}
        \multirow{2}{*}{\bf Metric}& \multirow{2}{*}{\bf Encoder}& \multicolumn{3}{c}{\textbf{fl}} & \multicolumn{2}{|c|}{\textbf{non-fl}} & \multirow{2}{*}{\bf TB} \\
        & &    abs-idx &    rel-idx &  PoS-idx &     1p &    2p &   \\
        \hline
        \hline
        \multirow{16}{*}{\bf UAS}
            & LSTM &  58.8 &  60.5 &  61.5 &  74.4 &  73.3 &  70.4 \\
            &     &  75.5 &  78.3 &  77.4 &  82.5 &  81.7 &  79.0 \\
            &     &  80.1 &  82.6 &  80.3 &  83.9 &  82.6 &  78.6 \\
            & mGPT &  49.5 &  56.4 &  59.0 &  74.0 &  74.7 &  72.9 \\
            &     &  73.9 &  78.2 &  81.9 &  88.1 &  87.4 &  83.3 \\
            &     &  78.6 &  83.4 &  85.6 &  88.7 &  88.6 &  85.2 \\
            & BLOOM &  44.4 &  54.8 &  59.0 &  73.2 &  72.8 &  69.0 \\
            &     &  68.7 &  76.3 &  80.4 &  85.3 &  84.6 &  81.7 \\
            &     &  74.0 &  80.8 &  82.2 &  85.7 &  86.4 &  81.3 \\
            & BiLSTM &  82.4 &  88.9 &  77.4 &  87.1 &  86.0 &  83.5 \\
            &     &  84.6 &  87.3 &  84.8 &  84.6 &  83.7 &  80.5 \\
            &     &  84.7 &  86.8 &  85.1 &  84.7 &  84.0 &  79.1 \\
            & XLM &  88.6 &  90.1 &  78.3 &  91.6 &  91.0 &  89.5 \\
            &     &  89.9 &  92.9 &  90.7 &  92.6 &  91.2 &  89.4 \\
            &     &  92.2 &  92.5 &  92.6 &  93.1 &  92.8 &  86.4 \\
        \cline{2-8}
            &\bf DM & \multicolumn{6}{c}{90.6}\\
        \hline 
        \multirow{16}{*}{\bf LAS}
            & LSTM &  55.0 &  56.8 &  59.0 &  67.5 &  66.3 &  60.5 \\
            &     &  71.3 &  73.8 &  73.9 &  77.7 &  77.1 &  62.9 \\
            &     &  76.0 &  78.5 &  76.6 &  79.4 &  78.6 &  64.6 \\
            & mGPT &  46.0 &  52.3 &  55.7 &  65.7 &  66.2 &  65.4 \\
            &     &  70.9 &  74.9 &  79.1 &  84.2 &  83.5 &  83.3 \\
            &     &  75.4 &  80.2 &  83.0 &  85.0 &  84.8 &  85.2 \\
            & BLOOM &  41.0 &  50.2 &  55.1 &  64.0 &  63.9 &  58.3 \\
            &     &  64.1 &  72.4 &  76.2 &  80.7 &  79.4 &  68.4 \\
            &     &  70.4 &  76.3 &  78.6 &  81.4 &  81.9 &  70.6 \\
            & BiLSTM &  80.1 &  86.6 &  75.7 &  84.9 &  83.6 &  76.2 \\
            &     &  80.7 &  83.6 &  81.3 &  80.5 &  79.7 &  66.8 \\
            &     &  80.7 &  82.9 &  81.7 &  81.0 &  80.1 &  65.4 \\
            & XLM &  86.5 &  87.9 &  76.4 &  89.1 &  88.5 &  84.5 \\
            &     &  86.8 &  90.5 &  88.7 &  90.2 &  88.6 &  74.4 \\
            &     &  89.9 &  90.1 &  90.4 &  90.9 &  90.5 &  74.2 \\
        \cline{2-8}
        & \bf DM & \multicolumn{6}{c}{88.5}\\
\end{tabular}
    \caption{\label{english-results}UAS and LAS  for the English EWT treebank. Notation comes from Table \ref{arabic-results}.} 
\end{table}

%% file: results/french-gsd/complete-delay.tex
\begin{table}[bpth]
    \centering
    \scriptsize
    \addtolength{\tabcolsep}{-3pt}
    \def\arraystretch{1.3}
    \begin{tabular}{c|c|ccc|cc|c}
        \multirow{2}{*}{\bf Metric}& \multirow{2}{*}{\bf Encoder}& \multicolumn{3}{c}{\textbf{fl}} & \multicolumn{2}{|c|}{\textbf{non-fl}} & \multirow{2}{*}{\bf TB} \\
        & &    abs-idx &    rel-idx &  PoS-idx &     1p &    2p &   \\
        \hline
        \hline
        \multirow{16}{*}{\bf UAS}
            & LSTM &  69.2 &  70.9 &  67.0 &  82.2 &  81.6 &  84.7 \\
            &     &  77.5 &  79.4 &  80.3 &  80.9 &  81.6 &  84.4 \\
            &     &  81.0 &  83.3 &  83.1 &  84.5 &  84.5 &  84.3 \\
            & mGPT &  57.9 &  68.3 &  67.4 &  84.4 &  84.2 &  84.1 \\
            &     &  73.8 &  79.8 &  84.4 &  85.8 &  86.2 &  87.5 \\
            &     &  78.0 &  83.6 &  87.6 &  89.2 &  88.8 &  86.3 \\
            & BLOOM &  54.1 &  68.0 &  66.1 &  82.8 &  82.2 &  83.8 \\
            &     &  69.2 &  78.5 &  82.3 &  85.0 &  84.7 &  85.8 \\
            &     &  72.0 &  82.8 &  86.7 &  87.0 &  86.8 &  86.7 \\
            & BiLSTM &  85.3 &  91.4 &  80.1 &  89.8 &  89.0 &  88.8 \\
            &     &  84.7 &  88.4 &  87.6 &  87.6 &  86.7 &  84.6 \\
            &     &  85.0 &  87.8 &  87.6 &  87.2 &  86.8 &  85.5 \\
            & XLM &  92.3 &  92.2 &  80.7 &  94.5 &  94.7 &  91.6 \\
            &     &  93.0 &  93.4 &  93.5 &  93.8 &  94.4 &  90.2 \\
            &     &  94.4 &  93.8 &  92.4 &  93.5 &  93.9 &  89.7 \\
        \cline{2-8}
            &\bf DM & \multicolumn{6}{c}{93.3}\\
        \hline 
        \multirow{16}{*}{\bf LAS}
            & LSTM &  65.4 &  67.1 &  63.4 &  75.8 &  75.3 &  76.0 \\
            &     &  72.1 &  74.3 &  75.1 &  75.1 &  75.5 &  66.6 \\
            &     &  75.5 &  78.2 &  78.3 &  79.0 &  79.0 &  68.6 \\
            & mGPT &  54.9 &  64.7 &  63.8 &  78.2 &  78.2 &  78.1 \\
            &     &  70.6 &  76.4 &  80.9 &  81.4 &  81.9 &  76.0 \\
            &     &  74.9 &  80.6 &  84.2 &  85.4 &  85.3 &  75.9 \\
            & BLOOM &  50.6 &  63.9 &  62.5 &  76.4 &  75.6 &  73.6 \\
            &     &  65.7 &  74.4 &  78.7 &  79.8 &  79.7 &  73.3 \\
            &     &  68.5 &  78.8 &  83.2 &  82.0 &  82.0 &  74.4 \\
            & BiLSTM &  82.0 &  87.7 &  77.2 &  86.2 &  85.5 &  82.0 \\
            &     &  79.4 &  83.2 &  82.5 &  82.4 &  81.6 &  71.1 \\
            &     &  79.7 &  82.2 &  82.5 &  82.1 &  81.5 &  71.9 \\
            & XLM &  90.0 &  89.9 &  79.1 &  92.0 &  92.4 &  87.0 \\
            &     &  90.2 &  90.7 &  90.7 &  91.1 &  91.7 &  79.8 \\
            &     &  91.6 &  91.0 &  89.8 &  90.9 &  90.9 &  78.4 \\
        \cline{2-8}
        & \bf DM & \multicolumn{6}{c}{89.6}\\
\end{tabular}
    \caption{\label{french-results}UAS and LAS  for the French GSD treebank. Notation comes from Table \ref{arabic-results}.} 
\end{table}

%% file: results/hindi-hdtb/complete-delay.tex
\begin{table}[bpth]
    \centering
    \scriptsize
    \addtolength{\tabcolsep}{-3pt}
    \def\arraystretch{1.3}
    \begin{tabular}{c|c|ccc|cc|c}
        \multirow{2}{*}{\bf Metric}& \multirow{2}{*}{\bf Encoder}& \multicolumn{3}{c}{\textbf{fl}} & \multicolumn{2}{|c|}{\textbf{non-fl}} & \multirow{2}{*}{\bf TB} \\
        & &    abs-idx &    rel-idx &  PoS-idx &     1p &    2p &   \\
        \hline
        \hline
        \multirow{16}{*}{\bf UAS}
            & LSTM &  66.0 &  66.1 &  67.1 &  82.8 &  82.7 &  83.2 \\
            &     &  73.0 &  74.5 &  81.4 &  81.1 &  80.8 &  87.6 \\
            &     &  76.8 &  78.7 &  84.1 &  86.0 &  86.1 &  88.9 \\
            & mGPT &  46.2 &  62.4 &  66.1 &  83.5 &  83.4 &  82.3 \\
            &     &  67.7 &  72.4 &  82.5 &  86.2 &  85.8 &  88.7 \\
            &     &  73.0 &  77.6 &  85.8 &  90.1 &  89.6 &  88.1 \\
            & BLOOM &  48.0 &  61.6 &  64.9 &  80.8 &  81.2 &  81.7 \\
            &     &  61.9 &  70.3 &  81.9 &  84.2 &  83.9 &  88.3 \\
            &     &  64.7 &  75.1 &  85.1 &  88.2 &  88.0 &  88.8 \\
            & BiLSTM &  91.5 &  94.2 &  78.2 &  93.3 &  93.3 &  92.8 \\
            &     &  87.8 &  91.2 &  90.4 &  88.2 &  87.8 &  90.7 \\
            &     &  88.1 &  91.4 &  91.1 &  87.5 &  87.0 &  89.6 \\
            & XLM &  94.3 &  93.5 &  78.0 &  94.7 &  95.3 &  93.8 \\
            &     &  94.6 &  94.1 &  92.7 &  93.7 &  94.2 &  94.1 \\
            &     &  95.0 &  94.2 &  92.6 &  94.0 &  94.2 &  93.8 \\
        \cline{2-8}
            &\bf DM & \multicolumn{6}{c}{95.5}\\
        \hline 
        \multirow{16}{*}{\bf LAS}
            & LSTM &  60.0 &  60.4 &  60.7 &  67.8 &  67.8 &  71.3 \\
            &     &  67.8 &  69.4 &  76.3 &  74.1 &  73.8 &  72.0 \\
            &     &  72.4 &  74.2 &  79.4 &  80.3 &  80.4 &  74.4 \\
            & mGPT &  41.5 &  57.0 &  60.5 &  69.2 &  69.6 &  73.1 \\
            &     &  64.0 &  68.5 &  78.1 &  79.6 &  79.1 &  77.4 \\
            &     &  69.2 &  73.8 &  82.0 &  84.9 &  84.1 &  77.4 \\
            & BLOOM &  43.4 &  55.0 &  57.5 &  65.4 &  66.0 &  69.2 \\
            &     &  57.6 &  66.1 &  77.6 &  77.1 &  77.1 &  74.5 \\
            &     &  61.0 &  71.0 &  80.7 &  82.7 &  82.2 &  74.1 \\
            & BiLSTM &  88.5 &  91.2 &  75.6 &  90.4 &  90.3 &  83.8 \\
            &     &  83.0 &  86.4 &  85.9 &  83.7 &  83.2 &  75.5 \\
            &     &  83.4 &  86.6 &  86.7 &  82.8 &  82.2 &  75.2 \\
            & XLM &  91.3 &  90.5 &  75.7 &  91.8 &  92.3 &  86.0 \\
            &     &  91.3 &  90.9 &  89.8 &  90.3 &  90.8 &  85.8 \\
            &     &  91.8 &  90.8 &  89.7 &  90.5 &  90.8 &  88.2 \\
        \cline{2-8}
        & \bf DM & \multicolumn{6}{c}{92.7}\\
\end{tabular}
    \caption{\label{hindi-results}UAS and LAS  for the Hindi HDTB treebank. Notation comes from Table \ref{arabic-results}.} 
\end{table}

%% file: results/indonesian-gsd/complete-delay.tex
\begin{table}[bpth]
    \centering
    \scriptsize
    \addtolength{\tabcolsep}{-3pt}
    \def\arraystretch{1.3}
    \begin{tabular}{c|c|ccc|cc|c}
        \multirow{2}{*}{\bf Metric}& \multirow{2}{*}{\bf Encoder}& \multicolumn{3}{c}{\textbf{fl}} & \multicolumn{2}{|c|}{\textbf{non-fl}} & \multirow{2}{*}{\bf TB} \\
        & &    abs-idx &    rel-idx &  PoS-idx &     1p &    2p &   \\
        \hline
        \hline
        \multirow{16}{*}{\bf UAS}
            & LSTM &  70.4 &  73.0 &  57.5 &  77.9 &  77.3 &  78.6 \\
            &     &  78.3 &  79.2 &  80.6 &  77.6 &  76.4 &  79.2 \\
            &     &  75.9 &  80.2 &  81.4 &  79.7 &  80.0 &  79.5 \\
            & mGPT &  46.0 &  67.5 &  56.0 &  77.2 &  77.0 &  76.0 \\
            &     &  60.6 &  75.5 &  80.8 &  80.4 &  80.2 &  78.8 \\
            &     &  62.0 &  77.7 &  81.9 &  82.0 &  82.1 &  79.6 \\
            & BLOOM &  40.4 &  66.2 &  55.9 &  75.9 &  75.9 &  78.4 \\
            &     &  55.8 &  73.6 &  80.6 &  78.0 &  77.6 &  77.2 \\
            &     &  56.0 &  76.2 &  80.5 &  80.1 &  80.4 &  77.9 \\
            & BiLSTM &  73.9 &  84.5 &  67.6 &  81.5 &  83.8 &  84.7 \\
            &     &  79.0 &  83.6 &  82.4 &  82.2 &  79.9 &  78.3 \\
            &     &  76.5 &  82.5 &  82.7 &  80.9 &  80.8 &  80.1 \\
            & XLM &  75.5 &  82.9 &  65.4 &  86.2 &  86.3 &  85.1 \\
            &     &  83.7 &  86.0 &  87.2 &  87.5 &  87.1 &  83.0 \\
            &     &  83.4 &  85.6 &  84.6 &  87.6 &  87.2 &  83.2 \\
        \cline{2-8}
            &\bf DM & \multicolumn{6}{c}{88.6}\\
        \hline 
        \multirow{16}{*}{\bf LAS}
            & LSTM &  65.7 &  68.2 &  53.8 &  72.1 &  71.6 &  67.7 \\
            &     &  69.9 &  70.8 &  72.3 &  69.2 &  68.8 &  59.4 \\
            &     &  68.6 &  72.4 &  73.4 &  72.0 &  71.9 &  59.7 \\
            & mGPT &  40.9 &  60.1 &  50.6 &  67.3 &  67.6 &  64.9 \\
            &     &  54.8 &  68.5 &  73.6 &  72.3 &  71.8 &  66.9 \\
            &     &  56.5 &  70.7 &  75.2 &  74.2 &  74.6 &  68.3 \\
            & BLOOM &  35.8 &  58.6 &  49.7 &  66.2 &  66.1 &  63.3 \\
            &     &  50.0 &  66.1 &  73.2 &  69.3 &  68.9 &  64.3 \\
            &     &  50.4 &  68.5 &  73.4 &  72.0 &  71.8 &  65.8 \\
            & BiLSTM &  69.7 &  79.5 &  64.8 &  73.4 &  78.8 &  73.5 \\
            &     &  71.0 &  75.3 &  74.1 &  73.3 &  71.7 &  61.9 \\
            &     &  68.7 &  74.1 &  75.0 &  72.4 &  72.6 &  62.4 \\
            & XLM &  69.5 &  76.1 &  61.5 &  78.8 &  79.2 &  71.7 \\
            &     &  77.3 &  78.8 &  80.6 &  80.3 &  79.7 &  71.1 \\
            &     &  76.9 &  78.8 &  79.0 &  80.8 &  80.3 &  70.7 \\
        \cline{2-8}
        & \bf DM & \multicolumn{6}{c}{83.6}\\
\end{tabular}
    \caption{\label{indonesian-results}UAS and LAS  for the Indonesian GSD treebank. Notation comes from Table \ref{arabic-results}.} 
\end{table}

%% file: results/marathi-ufal/complete-delay.tex
\begin{table}[bpth]
    \centering
    \scriptsize
    \addtolength{\tabcolsep}{-3pt}
    \def\arraystretch{1.3}
    \begin{tabular}{c|c|ccc|cc|c}
        \multirow{2}{*}{\bf Metric}& \multirow{2}{*}{\bf Encoder}& \multicolumn{3}{c}{\textbf{fl}} & \multicolumn{2}{|c|}{\textbf{non-fl}} & \multirow{2}{*}{\bf TB} \\
        & &    abs-idx &    rel-idx &  PoS-idx &     1p &    2p &   \\
        \hline
        \hline
        \multirow{16}{*}{\bf UAS}
            & LSTM &  42.7 &  55.3 &  64.1 &  65.0 &  69.7 &  65.8 \\
            &     &  54.7 &  63.0 &  65.2 &  58.4 &  62.8 &  69.9 \\
            &     &  60.8 &  64.2 &  60.6 &  63.1 &  60.1 &  71.1 \\
            & mGPT &  27.2 &  45.2 &  59.2 &  63.1 &  61.2 &  60.7 \\
            &     &  39.2 &  58.2 &  66.8 &  65.8 &  61.5 &  72.7 \\
            &     &  37.4 &  58.1 &  68.9 &  63.1 &  64.4 &  66.8 \\
            & BLOOM &  28.9 &  45.6 &  59.0 &  59.7 &  58.2 &  60.0 \\
            &     &  46.7 &  55.1 &  73.5 &  62.3 &  64.9 &  69.2 \\
            &     &  43.3 &  58.7 &  72.3 &  61.0 &  64.9 &  69.0 \\
            & BiLSTM &  60.4 &  75.7 &  69.7 &  75.2 &  73.8 &  76.0 \\
            &     &  60.2 &  69.3 &  67.8 &  63.9 &  61.6 &  66.6 \\
            &     &  58.8 &  66.3 &  65.8 &  55.5 &  65.1 &  67.4 \\
            & XLM &  33.7 &  63.4 &  70.4 &  71.4 &  71.8 &  73.5 \\
            &     &  54.5 &  69.7 &  78.6 &  72.1 &  73.6 &  79.7 \\
            &     &  52.8 &  74.7 &  79.4 &  80.0 &  68.0 &  80.2 \\
        \cline{2-8}
            &\bf DM & \multicolumn{6}{c}{82.4}\\
        \hline 
        \multirow{16}{*}{\bf LAS}
            & LSTM &  34.0 &  43.9 &  47.8 &  44.2 &  45.9 &  29.8 \\
            &     &  39.8 &  48.8 &  50.2 &  44.0 &  48.4 &  38.3 \\
            &     &  43.9 &  50.3 &  46.6 &  43.8 &  44.7 &  40.2 \\
            & mGPT &  21.8 &  34.2 &  48.3 &  45.9 &  45.6 &  36.4 \\
            &     &  31.5 &  45.3 &  50.9 &  53.0 &  48.0 &  72.7 \\
            &     &  27.4 &  45.5 &  55.4 &  49.9 &  51.2 &  66.8 \\
            & BLOOM &  23.8 &  34.5 &  48.3 &  44.4 &  41.5 &  28.2 \\
            &     &  36.8 &  44.7 &  58.2 &  49.8 &  54.4 &  42.4 \\
            &     &  33.7 &  46.6 &  57.4 &  49.9 &  49.2 &  42.8 \\
            & BiLSTM &  47.6 &  60.0 &  57.3 &  59.0 &  59.2 &  49.3 \\
            &     &  44.0 &  51.8 &  44.8 &  44.9 &  44.3 &  42.6 \\
            &     &  42.4 &  49.8 &  47.3 &  42.4 &  45.3 &  30.8 \\
            & XLM &  30.6 &  55.3 &  64.6 &  62.6 &  62.4 &  61.6 \\
            &     &  49.1 &  58.0 &  66.5 &  61.2 &  60.8 &  64.4 \\
            &     &  45.9 &  64.2 &  65.4 &  69.8 &  55.2 &  55.6 \\
        \cline{2-8}
        & \bf DM & \multicolumn{6}{c}{62.5}\\
\end{tabular}
    \caption{\label{marathi-results}UAS and LAS  for the Marathi UFAL treebank. Notation comes from Table \ref{arabic-results}.} 
\end{table}

%% file: results/spanish-ancora/complete-delay.tex
\begin{table}[bpth]
    \centering
    \scriptsize
    \addtolength{\tabcolsep}{-3pt}
    \def\arraystretch{1.3}
    \begin{tabular}{c|c|ccc|cc|c}
        \multirow{2}{*}{\bf Metric}& \multirow{2}{*}{\bf Encoder}& \multicolumn{3}{c}{\textbf{fl}} & \multicolumn{2}{|c|}{\textbf{non-fl}} & \multirow{2}{*}{\bf TB} \\
        & &    abs-idx &    rel-idx &  PoS-idx &     1p &    2p &   \\
        \hline
        \hline
        \multirow{16}{*}{\bf UAS}
            & LSTM &  67.4 &  67.9 &  64.7 &  80.4 &  79.9 &  82.3 \\
            &     &  75.9 &  78.6 &  80.6 &  79.2 &  78.3 &  82.8 \\
            &     &  77.7 &  82.0 &  84.0 &  82.1 &  82.0 &  83.5 \\
            & mGPT &  62.0 &  66.4 &  65.4 &  82.4 &  83.2 &  82.9 \\
            &     &  73.0 &  79.8 &  84.9 &  85.4 &  84.8 &  85.2 \\
            &     &  76.2 &  83.5 &  88.4 &  87.6 &  87.2 &  85.6 \\
            & BLOOM &  53.1 &  66.7 &  65.1 &  81.0 &  81.4 &  81.7 \\
            &     &  67.7 &  78.4 &  83.9 &  83.7 &  83.6 &  84.2 \\
            &     &  70.3 &  82.3 &  87.1 &  85.7 &  85.8 &  85.2 \\
            & BiLSTM &  88.4 &  89.6 &  79.1 &  88.5 &  87.4 &  88.0 \\
            &     &  84.8 &  87.4 &  88.2 &  86.3 &  85.9 &  84.5 \\
            &     &  87.8 &  87.1 &  87.6 &  85.8 &  85.8 &  83.8 \\
            & XLM &  92.0 &  91.5 &  80.5 &  93.4 &  93.4 &  91.2 \\
            &     &  93.8 &  93.0 &  94.2 &  93.6 &  93.4 &  88.7 \\
            &     &  93.9 &  93.0 &  92.7 &  93.6 &  93.6 &  88.6 \\
        \cline{2-8}
            &\bf DM & \multicolumn{6}{c}{93.1}\\
        \hline 
        \multirow{16}{*}{\bf LAS}
            & LSTM &  63.7 &  63.8 &  61.6 &  75.3 &  74.6 &  73.5 \\
            &     &  71.2 &  74.0 &  76.1 &  74.4 &  73.8 &  67.0 \\
            &     &  73.1 &  77.3 &  79.8 &  77.8 &  77.6 &  68.5 \\
            & mGPT &  59.3 &  63.1 &  62.9 &  78.0 &  78.9 &  76.3 \\
            &     &  70.5 &  77.2 &  82.4 &  82.7 &  82.0 &  85.2 \\
            &     &  73.6 &  80.9 &  86.2 &  85.0 &  84.5 &  85.6 \\
            & BLOOM &  50.3 &  63.2 &  62.2 &  76.0 &  76.2 &  74.5 \\
            &     &  65.1 &  75.3 &  81.0 &  80.4 &  80.3 &  72.3 \\
            &     &  67.5 &  79.2 &  84.6 &  82.5 &  82.2 &  74.2 \\
            & BiLSTM &  84.9 &  86.2 &  76.3 &  85.3 &  84.0 &  80.9 \\
            &     &  80.4 &  83.1 &  84.1 &  82.1 &  81.7 &  71.3 \\
            &     &  82.9 &  82.7 &  83.4 &  81.7 &  81.5 &  71.2 \\
            & XLM &  90.0 &  89.7 &  79.2 &  91.4 &  91.6 &  85.5 \\
            &     &  91.6 &  91.0 &  92.3 &  91.6 &  91.4 &  78.2 \\
            &     &  91.7 &  90.8 &  90.9 &  91.6 &  91.5 &  76.4 \\
        \cline{2-8}
        & \bf DM & \multicolumn{6}{c}{82.4}\\
\end{tabular}
    \caption{\label{spanish-results}UAS and LAS  for the Spanish ANCORA treebank. Notation comes from Table \ref{arabic-results}.} 
\end{table}

%% file: results/tamil-ttb/complete-delay.tex
\begin{table}[bpth]
    \centering
    \scriptsize
    \addtolength{\tabcolsep}{-3pt}
    \def\arraystretch{1.3}
    \begin{tabular}{c|c|ccc|cc|c}
        \multirow{2}{*}{\bf Metric}& \multirow{2}{*}{\bf Encoder}& \multicolumn{3}{c}{\textbf{fl}} & \multicolumn{2}{|c|}{\textbf{non-fl}} & \multirow{2}{*}{\bf TB} \\
        & &    abs-idx &    rel-idx &  PoS-idx &     1p &    2p &   \\
        \hline
        \hline
        \multirow{16}{*}{\bf UAS}
            & LSTM &  39.4 &  55.7 &  59.1 &  67.3 &  66.0 &  69.7 \\
            &     &  42.5 &  50.7 &  54.3 &  54.4 &  43.5 &  61.7 \\
            &     &  46.5 &  57.1 &  56.0 &  53.7 &  50.8 &  61.2 \\
            & mGPT &  20.3 &  48.7 &  55.8 &  64.1 &  62.8 &  65.2 \\
            &     &  32.1 &  57.7 &  68.3 &  62.0 &  60.3 &  67.0 \\
            &     &  28.7 &  60.9 &  69.9 &  54.1 &  55.1 &  66.9 \\
            & BLOOM &  19.4 &  49.4 &  54.0 &  55.8 &  63.1 &  65.7 \\
            &     &  28.6 &  56.0 &  68.5 &  62.4 &  58.8 &  66.7 \\
            &     &  31.6 &  60.0 &  64.9 &  64.9 &  63.6 &  71.9 \\
            & BiLSTM &  51.2 &  71.4 &  61.7 &  74.3 &  74.7 &  73.7 \\
            &     &  51.8 &  57.0 &  63.3 &  57.6 &  57.8 &  64.0 \\
            &     &  45.1 &  57.3 &  64.2 &  58.2 &  56.4 &  65.6 \\
            & XLM &  28.9 &  65.9 &  62.4 &  75.8 &  74.6 &  78.6 \\
            &     &  41.0 &  69.6 &  72.2 &  71.8 &  72.4 &  74.4 \\
            &     &  39.7 &  68.3 &  75.5 &  74.3 &  71.2 &  75.6 \\
        \cline{2-8}
            &\bf DM & \multicolumn{6}{c}{75.8}\\
        \hline 
        \multirow{16}{*}{\bf LAS}
            & LSTM &  33.2 &  44.8 &  50.6 &  52.6 &  51.3 &  41.0 \\
            &     &  28.3 &  35.6 &  40.3 &  38.6 &  22.5 &  32.4 \\
            &     &  30.7 &  40.1 &  42.4 &  38.3 &  36.7 &  29.6 \\
            & mGPT &  15.7 &  37.6 &  44.6 &  47.5 &  46.5 &  42.7 \\
            &     &  25.8 &  47.4 &  56.4 &  50.3 &  48.4 &  67.0 \\
            &     &  22.8 &  49.5 &  58.2 &  41.6 &  43.1 &  44.7 \\
            & BLOOM &  15.4 &  37.2 &  42.9 &  38.6 &  46.8 &  40.3 \\
            &     &  22.7 &  45.0 &  56.2 &  50.1 &  47.2 &  45.2 \\
            &     &  25.9 &  48.5 &  52.1 &  53.8 &  50.3 &  51.8 \\
            & BiLSTM &  44.9 &  62.3 &  47.0 &  65.7 &  66.3 &  57.5 \\
            &     &  38.0 &  40.9 &  45.6 &  41.1 &  42.9 &  32.5 \\
            &     &  31.5 &  39.5 &  47.8 &  41.1 &  41.4 &  35.4 \\
            & XLM &  24.9 &  56.8 &  55.0 &  65.6 &  64.7 &  64.8 \\
            &     &  34.7 &  59.4 &  61.7 &  60.8 &  61.7 &  53.2 \\
            &     &  33.2 &  58.0 &  65.2 &  63.7 &  61.2 &  55.7 \\
        \cline{2-8}
        & \bf DM & \multicolumn{6}{c}{65.5}\\
\end{tabular}
    \caption{\label{tamil-results}UAS and LAS  for the Tamil TTB treebank. Notation comes from Table \ref{arabic-results}.} 
\end{table}

%% file: results/telugu-mtg/complete-results.tex
\begin{table}[bpth]
    \centering
    \scriptsize
    \addtolength{\tabcolsep}{-3pt}
    \def\arraystretch{1.3}
    \begin{tabular}{c|c|ccc|cc|c}
        \multirow{2}{*}{\bf Metric}& \multirow{2}{*}{\bf Encoder}& \multicolumn{3}{c}{\textbf{fl}} & \multicolumn{2}{|c|}{\textbf{non-fl}} & \multirow{2}{*}{\bf TB} \\
        & &    abs-idx &    rel-idx &  PoS-idx &     1p &    2p &   \\
        \hline
        \hline
        \multirow{16}{*}{\bf UAS}
            & LSTM &  64.6 &  67.6 &  73.8 &  84.7 &  85.0 &  80.4 \\
            &     &  78.6 &  78.3 &  78.4 &  82.2 &  83.5 &  87.3 \\
            &     &  78.3 &  83.1 &  82.5 &  83.9 &  85.0 &  90.0 \\
            & mGPT &  55.5 &  59.1 &  71.8 &  82.0 &  81.6 &  75.6 \\
            &     &  73.7 &  75.0 &  85.0 &  84.5 &  85.0 &  87.4 \\
            &     &  80.2 &  82.7 &  86.7 &  89.6 &  84.2 &  80.8 \\
            & BLOOM &  51.7 &  54.8 &  70.0 &  79.1 &  79.2 &  74.1 \\
            &     &  72.4 &  73.6 &  84.3 &  84.5 &  82.9 &  87.2 \\
            &     &  78.7 &  78.2 &  85.7 &  84.0 &  84.0 &  87.8 \\
            & BiLSTM &  86.3 &  89.9 &  85.3 &  90.2 &  90.3 &  89.0 \\
            &     &  71.8 &  88.1 &  86.7 &  80.0 &  84.5 &  81.0 \\
            &     &  71.3 &  88.1 &  85.0 &  83.2 &  76.2 &  82.4 \\
            & XLM &  77.8 &  87.4 &  84.9 &  89.6 &  89.2 &  88.9 \\
            &     &  83.6 &  88.2 &  89.9 &  86.7 &  87.0 &  84.2 \\
            &     &  68.6 &  90.7 &  91.5 &  89.0 &  88.3 &  91.7 \\
        \cline{2-8}
            &\bf DM & \multicolumn{6}{c}{89.7}\\
        \hline 
        \multirow{16}{*}{\bf LAS}
            & LSTM &  57.8 &  59.9 &  64.8 &  66.6 &  67.0 &  53.5 \\
            &     &  68.6 &  67.0 &  67.0 &  62.9 &  63.2 &  52.3 \\
            &     &  47.7 &  70.6 &  69.7 &  60.8 &  67.6 &  48.0 \\
            & mGPT &  49.4 &  51.6 &  65.6 &  67.3 &  67.1 &  50.5 \\
            &     &  65.6 &  67.1 &  76.5 &  74.1 &  71.3 &  87.4 \\
            &     &  71.7 &  74.1 &  76.8 &  79.4 &  74.8 &  25.4 \\
            & BLOOM &  46.5 &  45.4 &  62.1 &  63.8 &  63.4 &  43.4 \\
            &     &  66.0 &  64.8 &  74.2 &  72.2 &  71.7 &  54.3 \\
            &     &  71.8 &  68.7 &  74.3 &  74.2 &  67.4 &  47.9 \\
            & BiLSTM &  74.9 &  77.7 &  76.7 &  79.6 &  79.6 &  64.6 \\
            &     &  52.2 &  70.4 &  70.5 &  60.2 &  64.9 &  47.1 \\
            &     &  52.5 &  73.9 &  64.9 &  60.8 &  55.0 &  56.5 \\
            & XLM &  72.3 &  79.8 &  79.5 &  81.7 &  81.1 &  61.3 \\
            &     &  75.5 &  79.2 &  80.4 &  71.2 &  67.5 &  35.7 \\
            &     &  51.9 &  82.2 &  83.6 &  77.2 &  78.9 &  55.9 \\
        \cline{2-8}
        & \bf DM & \multicolumn{6}{c}{61.8}\\
\end{tabular}
    \caption{\label{telugu-results}UAS and LAS  for the Telugu MTG treebank. Notation comes from Table \ref{arabic-results}.} 
\end{table}

%% file: results/vietnamese-vtb/complete-delay.tex
\begin{table}[bpth]
    \centering
    \scriptsize
    \addtolength{\tabcolsep}{-3pt}
    \def\arraystretch{1.3}
    \begin{tabular}{c|c|ccc|cc|c}
        \multirow{2}{*}{\bf Metric}& \multirow{2}{*}{\bf Encoder}& \multicolumn{3}{c}{\textbf{fl}} & \multicolumn{2}{|c|}{\textbf{non-fl}} & \multirow{2}{*}{\bf TB} \\
        & &    abs-idx &    rel-idx &  PoS-idx &     1p &    2p &   \\
        \hline
        \hline
        \multirow{16}{*}{\bf UAS}
            & LSTM &  52.8 &  57.3 &  53.9 &  63.2 &  64.5 &   64.5 \\
            &     &  49.2 &  57.8 &  58.2 &  57.2 &  56.7 &   60.9 \\
            &     &  50.2 &  59.8 &  60.0 &  59.6 &  57.4 &   61.9 \\
            & mGPT &  29.3 &  49.6 &  51.6 &  61.8 &  61.3 &   59.7 \\
            &     &  34.8 &  56.9 &  66.6 &  62.9 &  63.8 &   63.2 \\
            &     &  34.3 &  60.1 &  68.3 &  63.7 &  65.1 &   65.2 \\
            & BLOOM &  31.7 &  51.0 &  51.1 &  62.1 &  60.7 &   61.0 \\
            &     &  38.5 &  57.3 &  63.4 &  60.5 &  60.9 &   64.0 \\
            &     &  38.8 &  61.6 &  64.6 &  59.4 &  62.0 &   60.0 \\
            & BiLSTM &  66.8 &  72.7 &  64.0 &  70.1 &  70.5 &   71.0 \\
            &     &  57.5 &  64.6 &  62.9 &  60.0 &  61.1 &   61.8 \\
            &     &  56.2 &  64.3 &  62.5 &  60.2 &  60.0 &   60.9 \\
            & XLM &  48.1 &  68.8 &  64.4 &  74.1 &  73.8 &   76.0 \\
            &     &  58.1 &  68.6 &  73.2 &  73.4 &  71.4 &   75.3 \\
            &     &  58.2 &  68.1 &  72.6 &  70.8 &  69.8 &   75.0 \\
        \cline{2-8}
            &\bf DM & \multicolumn{6}{c}{76.6}\\
        \hline 
        \multirow{16}{*}{\bf LAS}
            & LSTM &  41.6 &  45.5 &  44.5 &  49.8 &  50.8 &   43.8 \\
            &     &  36.8 &  42.5 &  43.2 &  42.2 &  42.1 &   36.3 \\
            &     &  37.5 &  44.0 &  44.8 &  44.4 &  41.8 &   35.6 \\
            & mGPT &  23.3 &  36.7 &  41.2 &  47.4 &  46.7 &   39.8 \\
            &     &  28.1 &  44.3 &  53.6 &  49.4 &  50.2 &   63.2 \\
            &     &  26.8 &  47.0 &  54.4 &  49.9 &  51.2 &   45.9 \\
            & BLOOM &  24.9 &  38.2 &  40.5 &  46.7 &  45.9 &   39.8 \\
            &     &  30.2 &  44.3 &  50.6 &  46.4 &  47.5 &   44.3 \\
            &     &  30.7 &  48.1 &  51.0 &  45.8 &  48.6 &   37.4 \\
            & BiLSTM &  55.4 &  60.3 &  54.6 &  58.2 &  58.3 &   51.2 \\
            &     &  44.2 &  49.3 &  48.6 &  45.4 &  45.8 &   38.6 \\
            &     &  42.4 &  48.6 &  47.5 &  45.0 &  45.3 &   38.1 \\
            & XLM &  40.4 &  56.8 &  54.8 &  61.4 &  61.5 &   56.8 \\
            &     &  48.8 &  56.6 &  60.6 &  60.6 &  58.1 &   58.7\\
            &     &  48.8 &  55.8 &  60.0 &  57.1 &  56.6 &   57.8 \\
        \cline{2-8}
        & \bf DM & \multicolumn{6}{c}{61.9}\\
\end{tabular}
    \caption{\label{vietnamese-results}UAS and LAS  for the Vietnamese MTG treebank. Notation comes from Table \ref{arabic-results}.} 
\end{table}